\documentclass[lettersize,journal]{IEEEtran} 

\usepackage[nolist, nohyperlinks, printonlyused, withpage]{acronym}
\usepackage{adjustbox}
\usepackage{algorithm}
\usepackage{algpseudocode}
\usepackage{amsmath}
\usepackage{amssymb}  
\usepackage{booktabs}
\usepackage{cite}
\usepackage{hyperref}
\usepackage{graphicx}
\usepackage{tabularx}  
\usepackage{todonotes}
\usepackage[lofdepth,lotdepth,caption=false]{subfig}  
\usepackage{svg}

\usepackage{amsthm}  

\def\seqa{{\mathbf{a}}}
\def\seqb{{\mathbf{b}}}
\def\seqc{{\mathbf{c}}}

\def\seqe{{\mathbf{e}}}
\def\seqf{{\mathbf{f}}}
\def\seqg{{\mathbf{g}}}

\def\seqp{{\mathbf{p}}}

\def\seqr{{\mathbf{r}}}

\def\seqx{{\mathbf{x}}}

\def\eseqa{{\textnormal{a}}}
\def\eseqb{{\textnormal{b}}}
\def\eseqc{{\textnormal{c}}}

\def\eseqf{{\textnormal{f}}}
\def\eseqg{{\textnormal{g}}}

\def\eseqm{{\textnormal{m}}}

\newcommand{\isfalse}[0]{\text{false}}
\newcommand{\textif}[0]{, \text{if }}
\newcommand{\otherwise}[0]{, \text{otherwise}}

\newcommand{\cgt}[0]{{\succ}}

\newcommand{\abs}[1]{\left\vert#1\right\vert}
\newcommand{\countmod}[2]{\mathrm{count}\left(#1, #2 \right)}
\newcommand{\crowdingdistance}[1]{\mathrm{cd}\left(#1\right)}
\newcommand{\feasible}[1]{\mathrm{feasible}(#1)}

\newcommand{\len}[1]{\mathrm{len}(#1)}
\newcommand{\ndr}[1]{\mathrm{ndr}(#1)}
\newcommand{\motionplan}[1]{\bm{\xi}\left(#1\right)}
\newcommand{\norm}[1]{\left\lVert#1\right\rVert}

\newcommand{\size}[1]{\mathrm{size}(#1)}
\newcommand{\trace}[1]{\mathrm{Tr}(#1)}

\def\axisangle{\varphi}
\def\axistol{t_{\vu, \tolangle, \boldsymbol{\eps}}}

\def\constraints{\sC}

\def\destrans{\vt}
\def\desrot{\mR}
\def\eefpose{\seqp_{EEF}}
\def\euclideantol{t_{x, y, z}}
\def\isfalse{\mathtt{false}}

\def\goal{\seqg}
\def\goals{\sG}
\def\jointspace{\sQ}
\def\moduleset{\msM}

\def\nintegers{\left[ n \right]}
\def\morphology{\rvm}
\def\obstacles{\sO}
\def\poseerr{\seqe}
\def\pose{\seqp}

\def\projrot{d_{R}}
\def\projtrans{d_{t}}
\def\robot{\seqr}
\def\roterr{{\bm{R}_e}}
\def\tolangle{\vartheta}
\def\tol{t}

\def\path{\motionplan{t}}
\def\jointpath{\vq\left(t\right)}

\def\transerr{{\bm{t}_e}}
\def\istrue{\mathtt{true}}

\newcommand{\obstacleself}[2]{\obstacles_{\morphology, #2} \left( #1 \right)}

\DeclareMathOperator*{\mm}{mm}

\newcommand{\xhdr}[1]{\vspace{1.7mm}\noindent{{\bf #1.}}}

\theoremstyle{remark}

\newcommand{\figwdefault}{\linewidth}
\newcommand{\sixthfigurewidth}{\dimexpr 0.95\linewidth}

\usepackage{amsmath,amsfonts,bm}

\def\figref#1{Figure~\ref{#1}}
\def\Figref#1{Figure~\ref{#1}}

\let\oldeqref=\eqref
\let\eqref=\oldeqref  

\def\plaineqref#1{\oldeqref{#1}}

\def\1{\bm{1}}

\def\eps{{\epsilon}}

\def\rvm{{\mathbf{m}}}

\def\va{{\bm{a}}}
\def\vb{{\bm{b}}}
\def\vc{{\bm{c}}}

\def\vf{{\bm{f}}}

\def\vn{{\bm{n}}}

\def\vq{{\bm{q}}}

\def\vt{{\bm{t}}}
\def\vu{{\bm{u}}}

\def\evu{{u}}

\def\mA{{\bm{A}}}
\def\mB{{\bm{B}}}
\def\mC{{\bm{C}}}

\def\mR{{\bm{R}}}

\DeclareMathAlphabet{\mathsfit}{\encodingdefault}{\sfdefault}{m}{sl}
\SetMathAlphabet{\mathsfit}{bold}{\encodingdefault}{\sfdefault}{bx}{n}

\def\sA{{\mathbb{A}}}
\def\sB{{\mathbb{B}}}
\def\sC{{\mathbb{C}}}

\def\sF{{\mathbb{F}}}
\def\sG{{\mathbb{G}}}

\def\sM{{\mathbb{M}}}

\def\sO{{\mathbb{O}}}

\def\sQ{{\mathbb{Q}}}

\def\sS{{\mathbb{S}}}

\def\sX{{\mathbb{X}}}

\def\msA{{\mathcal{A}}}
\def\msB{{\mathcal{B}}}
\def\msC{{\mathcal{C}}}

\def\msM{{\mathcal{M}}}

\def\msX{{\mathcal{X}}}

\newcommand{\R}{\mathbb{R}}

\DeclareMathOperator*{\argmin}{arg\,min}

\title{\LARGE \bf
Holistic Construction Automation with Modular Robots:\\From High-Level Task Specification to Execution
}

\makeatletter
\author{
    Jonathan K\"ulz$^{1, 2}$,
    Michael Terzer$^{3}$,
    Marco Magri$^{3}$,
    Andrea Giusti$^{3}$,
    and Matthias Althoff$^{1, 2}$
\thanks{$^{1}$Technical University of Munich, Department of Computer Engineering}%
\thanks{$^{2}$Munich Center for Machine Learning (MCML)}%
\thanks{$^{3}$Fraunhofer Italia Research, Robotics and Intelligent Systems Engineering}%
\thanks{contact: jonathan.kuelz@tum.de}
}
\makeatother

\begin{document}
\bstctlcite{IEEEexample:BSTcontrol}  
\begin{acronym}
    \acro{bim}[BIM]{Building Information Modeling}
    \acro{modrob}[MRR]{modular reconfigurable robot}
    \acro{fk}[FK]{forward kinematics}
    \acro{ga}[GA]{genetic algorithm}
    \acro{ik}[IK]{inverse kinematics}
    \acro{nsga2}[NSGA-II]{nondominated sorting-based genetic algorithm}
    \acro{ompl}[OMPL]{Open Motion Planning Library }
    \acro{rl}[RL]{deep reinforcement learning}
    \acro{tcp}[TCP]{tool center point}
\end{acronym}
\maketitle
\thispagestyle{empty}
\pagestyle{empty}

\begin{abstract}
In situ robotic automation in construction is challenging due to constantly changing environments, a shortage of robotic experts, and a lack of standardized frameworks bridging robotics and construction practices.
This work proposes a holistic framework for construction task specification, optimization of robot morphology, and mission execution using a mobile modular reconfigurable robot.
Users can specify and monitor the desired robot behavior through a graphical interface.
In contrast to existing, monolithic solutions, we automatically identify a new task-tailored robot for every task by integrating \acf{bim}.
Our framework leverages modular robot components that enable the fast adaption of robot hardware to the specific demands of the construction task.
Other than previous works on modular robot optimization, we consider multiple competing objectives, which allow us to explicitly model the challenges of real-world transfer, such as calibration errors.
We demonstrate our framework in simulation by optimizing robots for drilling and spray painting.
Finally, experimental validation demonstrates that our approach robustly enables the autonomous execution of robotic drilling.

\end{abstract}

\def\abstractname{Note to Practitioners}
\begin{abstract}
Construction sites are unpredictable and constantly changing, making automation difficult. To tackle this, we propose a modular robotic system that can adapt to specific tasks.
In this work, we address the challenge of automating robotic drilling and introduce a general framework that also supports other construction tasks.
We use Building Information Modelling (BIM) data to automatically determine the best robot configuration by optimizing factors such as setup time and robustness to positioning errors.
Through an intuitive interface, workers can specify tasks, after which they assemble the robot following the provided instructions.

The result of this work is a robotic system that can be deployed on diverse construction sites while requiring only readily available BIM data and minimal user inputs.
Our approach enables the same hardware to be deployed across various scenarios while accounting for site-specific challenges. 
A core innovation lies in combining modular robotics with automated optimization, task planning, and motion planning to minimize the need for user intervention while delivering robots specifically tailored to each task.
To achieve this, we extend methods for modular robot design to account for multiple conflicting objectives in the optimization process.
Experimental results demonstrate the ability of the system to perform autonomous drilling with high precision.
Beyond drilling, this framework has the potential to support a wide range of applications, and we show its applicability to spray painting in simulation.
Overall, it offers a practical and adaptable solution for diverse construction tasks and sites.
\end{abstract}


\newpage
\section{Introduction}\label{sec:introduction}
\begin{figure*}
    \centering
    \includegraphics[width=\textwidth]{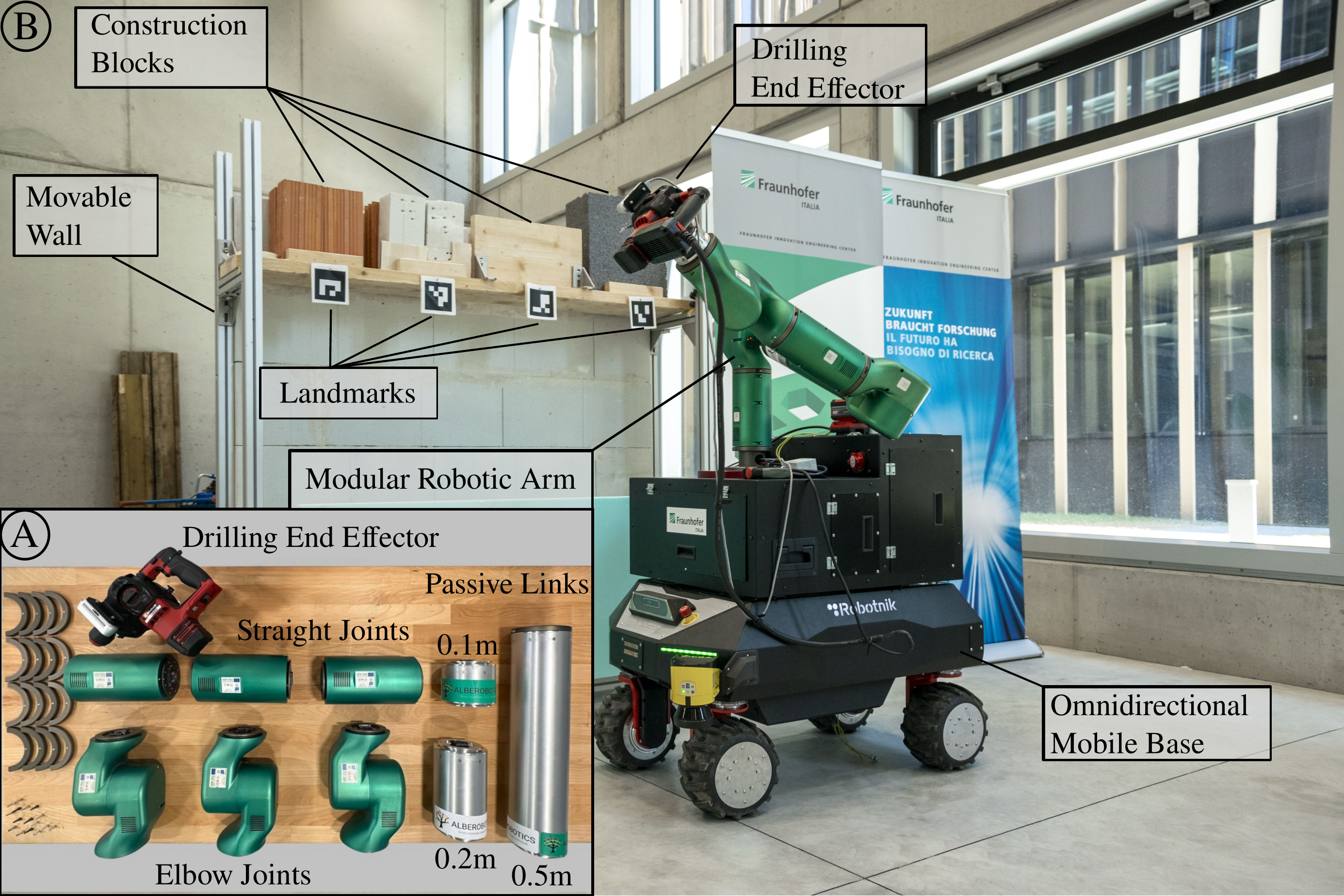}
    \caption{Section A shows the available robot modules for the manipulator, composed of six joints (green), three passive links (silver), and a custom end effector.
    Section B shows the assembled robotic arm on a mobile base and the experimental setup.}
    \label{fig:expSetup}
\end{figure*}
\IEEEPARstart{I}{n} recent decades, robotic automation in manufacturing significantly increased productivity~\cite{Graetz2018}.
To this day, a comparative technology leap in the construction sector has not occurred~\cite{Kim2015, Delgado2019, AutomationHandbookConstructionAutomation, Gappmaier2024}.
Furthermore, harsh and non-ergonomic working conditions result in labor shortages in the construction industry~\cite{Bock2015}.
Addressing these shortages through robotic automation requires versatility, adaptability, and robustness, as construction sites are complex, constantly changing, potentially hazardous, and challenging to navigate~\cite{Ardiny2015}.

In contrast to existing robotic systems for construction, the modular robot deployed in this work can easily be adapted to new tasks or changes in the environment in minutes.
The main contribution of this paper is a framework for in situ robotic automation that features (i) an autonomous decomposition of high-level instructions (e.g., ``drill holes'' or ``paint wall'') into executable skills, (ii) identification of an optimized module composition for the robot arm using \ac{bim} data of the environment, (iii) online motion planning, and (iv) adaptations to positioning errors of the mobile base.
The only actions required by the human worker are the specification of the task and the assembly of robot modules once the optimization of their composition is complete.
Although our framework is not limited to a particular task, for real-world evaluation, we focus on autonomous drilling as shown in~\Figref{fig:expSetup}.
A supplementary video of the real-world experiment and more can be found on the project website \mbox{\href{https://modular-construction-robot.cps.cit.tum.de}{modular-construction-robot.cps.cit.tum.de}}.

\section{Related Work}\label{sec:related_work}
This section discusses previous applications of mobile manipulators in construction and, specifically, drilling use cases.
In addition, we provide an overview of current methods for optimizing modular robot compositions.
We conclude with an overview of common strategies to tackle challenges in the transfer of simulation to real-world application.

\subsection{Robotic automation in building construction}
Some construction tasks, such as cutting bricks or insulation panels, can be automated by special machinery or in situ manufacturing processes~\cite{Slongo2024}.
However, highly accurate and flexible interactions with the environment require mobile robotic systems that can self-localize~\cite{Gawel2019}. 
Navigating the highly unstructured environments of construction sites challenges the sensing and perception capabilities of robots.
In~\cite{Doerfler2016}, 3D scans taken with onboard sensors enabled navigation and localization of a bricklaying robot in situ.
Perception can also be enriched with data from as-built \ac{bim}~\cite{Liao2023}.

The use of BIM data in construction has been shown to increase worker productivity~\cite{Poirier2015} and accelerate the deployment of construction robotics~\cite{Giusti2021}.
However, as reported by~\cite{Chen2018}, there is still a lack of research on integrating \ac{bim} with robotic systems in situ.
Previous work proposed the integration of BIM data into a mission parameterization toolchain for non-experts and validated its effectiveness through a spray-painting experiment~\cite{Terzer2024}. 

Particularly challenging use cases in construction robotics are precise power tool works on walls or ceilings, such as grinding or drilling holes.
The challenges stem from uncertainties in robot perception and disturbances induced by the operation of the power tools. 
Recent works present an aerial system for drilling holes limited to free-space areas and vertical walls or facades~\cite{Dautzenberg2023} or the development of a mobile manipulator with predefined reach and dexterity~\cite{Ortner2023}. 

Robots with fixed kinematics as presented in previous research~\cite{Helm2012, Outon2019, Stibinger2021}, may not provide the necessary flexibility required in construction~\cite{Pankert2022}.
This limitation is shared by most existing industrial solutions:
Robots such as ``Baubot'' or ``Hilti Jaibot'' can perform semi-autonomous drilling, but are limited to mostly unobstructed environments and predetermined operating heights.
Others, such as the ``Schindler R.I.S.E'', are tailored to specific environments, such as elevator shafts, but cannot be deployed anywhere else.
Reconfigurable modular robots -- comprising modular joints, links, and sensor modules -- can be adapted to specific requirements in situ while offering better transportability to and on the construction site. 
This aligns with previous research on digital in situ fabrication~\cite{Buchli2018}, highlighting that versatile, modular, and reusable solutions and real-world experiments are required to enable innovation in construction automation.

\subsection{Optimizing modular robot composition}
The optimization of robot morphologies, and more specifically, finding an optimal module composition for reconfigurable robots, has attracted attention for decades~\cite{Paredis1993, Han1997, Lipson2000, Yim2007, Althoff2019}.
Some approaches introduce heuristics and surrogate performance criteria to enable a brute-force search over reduced search spaces~\cite{Chen1998, Icer2016, Liu2020}.
Most approaches, however, use discrete black-box optimization algorithms~\cite{Icer2017, Whitman2020, Park2021}.
Since the early work of Karl Sims~\cite{Sims1994}, \acfp{ga} have been a common choice for the optimization of modular robot morphologies.
The work in~\cite{Romiti2023} deploys a \ac{ga} to co-optimize the morphology and base placement of a modular robot.
In ~\cite{Kuelz2024}, a genetic lexicographic algorithm is proposed to increase the efficiency and interpretability of the evolutionary optimization process.
While our work builds upon~\cite{Kuelz2024}, we extend it to multiobjective problems to explicitly model the trade-off between performance and robustness.
None of the above-mentioned works considers the challenges of deploying these robots outside idealized lab settings; the simulation is treated as a perfect representation of the world.
In contrast, we perform robot drilling using modular hardware while considering the challenges of simulation-to-reality transfer, such as inaccurate positioning and visual calibration of the robot arm.

\subsection{Robustness and multiobjective optimization}
It is well known that a purely simulation-based optimization of real-world systems can lead to suboptimal performance during deployment -- a phenomenon commonly referred to as the reality gap~\cite{Koos2010}.
This is particularly problematic in unstructured or partially structured environments, such as construction sites, where conditions are highly dynamic and unpredictable, and simulation often fails to capture the full complexity of real-world variability.
Especially in learning-based approaches, domain randomization has been successfully applied to address this issue~\cite{Tobin2017, Mordatch2015, Mehta2020}.
However, past work indicates a trade-off between the performance and generality of a solution~\cite{Andrychowicz2019, Fadini2022}.
In this work, we explicitly model this trade-off by formulating a multiobjective optimization problem with performance and robustness against positioning errors as competing objectives to find Pareto-optimal solutions.
To this end, we build on the \acf{nsga2}~\cite{Deb2002}.

\section{Framework}\label{sec:framework}
\begin{figure*}
    \centering
    \includegraphics[width=\figwdefault,trim=8 215 118 77,clip]{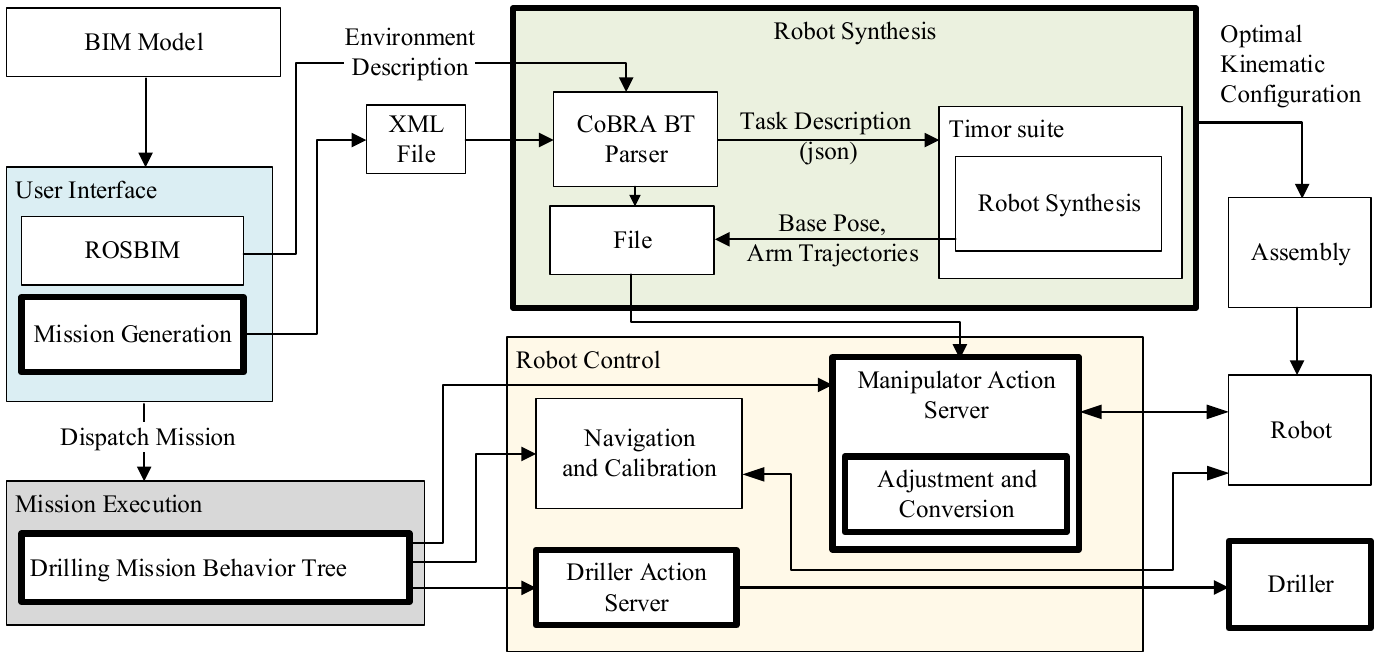}
    \caption{%
    The structure of the proposed framework from optimization of a robot model to mission execution. The bold boxes indicate the contributions of this work.
    During mission preparation, inputs from the user interface are used to perform robot synthesis.
    The mission execution and robot control block run during deployment.}
    \label{fig:sw_architecture}
\end{figure*}
Our framework is shown in \figref{fig:sw_architecture}.
It can be separated into four modules:
The user interface and robot synthesis modules are required for mission preparation, whereas mission execution and robot control run during deployment.
A user generates and parameterizes a mission through the BIM-based user interface presented in \cite{Terzer2024}.
Based on the user input, the robot synthesis module searches for Pareto-optimal configurations of the robot modules in simulation to execute the provided mission and proposes them to the user.
Next, the user assembles the modular robot and activates its control setup.
Finally, the robot performs visual calibration and executes the planned trajectory.

\subsection{Robot Synthesis}
Reconfigurable modular robot compositions are simulated and optimized using the toolbox for industrial modular robotics (Timor)~\cite{Kuelz2023}.
We rely on the modular robot modeling language introduced in CoBRA~\cite{Mayer2024} and a custom parser module (CoBRA BT Parser) to integrate the user interface and the \ac{bim} model using Timor.
After the optimization algorithm described in Section~\ref{sec:modrob_opt} identifies an optimized module composition and a motion trajectory, the worker can proceed with the physical robot assembly.
Self-identification of the robot~\cite{Romiti2022} or automatic model generation~\cite{Kuelz2023} can obtain a kinematic and dynamic robot model once its modules are provided.
The motion trajectory is stored and later passed to the robot control module.

\subsection{Robot control}
The robot control follows a modular setup.
While our framework is agnostic to the specific software that is used for control, we have integrated the following components for deployment:
\begin{itemize}
    \item Mobile base navigation, including local and global route planning, localization, and omnidirectional control, is based on the navigation2 framework~\cite{Macenski2020, macenski2024smac}.
    \item The manipulator control, including collision-free trajectory planning and execution, is based on Timor and joint impedance control using the manipulator control library\footnote{\url{https://github.com/FraunhoferItalia/fhi_manipulator_control_stack}}.
    \item The calibration module based on OpenCV\footnote{\url{https://github.com/opencv/opencv}} includes a pipeline that detects ArUco landmark poses~\cite{Jurado2016} with a camera to calibrate the robot position with respect to the environment.
    As the mobile base cannot move with arbitrary precision, even after calibration, the manipulator may not be exactly in the desired position.
    In this paper, we refer to this remaining error as the base positioning error.
\end{itemize}
\section{Modular Robot Composition Optimization}\label{sec:modrob_opt}
The modular nature of the robot realizes flexibly adapting its morphology.
However, it is a challenging task even for human experts to find a suitable or even optimized configuration for a specific task~\cite{Kuelz2023}.
In this section, we introduce an automatic optimization method for robot morphologies.
In contrast to previous work on robot design, we consider and jointly optimize multiple, potentially competing objectives $(a, b, \dots)$, such as cycle time, robustness, or the monetary cost of a solution.
Formally, the resulting constrained multiobjective optimization problem is given by
\begin{align}
    \max_{\eseqm_i \in \sM} & \Bigl( a(\morphology, \goals, \obstacles), b(\morphology, \goals, \obstacles), \dots \Bigr) \label{eq:formal_problem_def} \\
    \text{s.t. } & c_j(\morphology, \goals, \obstacles) = 0 \quad \text{for } j \in [m] \, , \nonumber
\end{align}
where $c_j$ are task constraints to be considered, and a sequence of modules fully defines the robot morphology \mbox{$\morphology = (\eseqm_1, \dots, \eseqm_i)$}.
A summary of the notation can be found in Table~\ref{tab:notation}.

Both constraints and objectives can be evaluated based on robot morphology $\morphology$, task-specific end-effector goals $\goals$, and obstacles $\obstacles$.
Without a clear hierarchy of objectives, \eqref{eq:formal_problem_def} generally does not have a single optimal solution.
Instead, our objective is to find all Pareto-optimal solutions.
To this end, we introduce a method based on lexicographic nondominant sorting optimization that combines the efficiency of lexicographic genetic optimizers with the versatility of the NSGA-II algorithm.
While lexicographic optimization is based on hierarchical ordering, we only use it to increase the efficiency of constraint evaluation for~\eqref{eq:formal_problem_def}.
No order is imposed on the optimization objectives $(a, b, \dots)$.

In Section~\ref{sec:preliminaries}, we include fundamental preliminaries, followed by formal definitions of our optimization approach, task definition, objective functions, and constraints in Sections~\ref{sec:lex_opt_method} to~\ref{sec:fitness_funs}.

\subsection{Preliminaries}\label{sec:preliminaries}
\begin{table}
\centering
\caption{Scores for exemplary robots and multiple objectives.}
    \begin{tabular}{@{}lllll@{}}
    \toprule
        & $\robot_1$ & $\robot_2$ & $\robot_3$ & $\robot_4$ \\
    \midrule
    $f_{fast}$  &   1   &   2   &   2   &   3 \\
    $f_{cheap}$ &   2   &   1   &   3   &   1 \\
    $f_{robust}$&   3   &   2   &   2   &   2 \\
    \bottomrule
    \end{tabular}
\label{tab:example_robots}
\end{table}
As a running example, consider a ranking of robots $\{1, 2, 3, 4\}$, based on how fast they can perform a specific task ($f_{fast}$), how cheap the corresponding hardware is ($f_{cheap}$), and how robust they are to external disturbances ($f_{robust}$).
Table~\ref{tab:example_robots} shows the scores $\robot_i$ of four exemplary robots.
The higher the score, the better the robot.
There are multiple objectives, so there is no clear ``best'' robot.
In the following sections, we will introduce a ranking-based selection method that considers multiple objectives.

\xhdr{Partially Ordered Set}
A partially ordered set is a tuple $(\sA, \leq)$ of a set $\sA$ and the order relation $\leq$, which is reflexive, antisymmetric, and transitive~\cite[Def. 1.1]{Schroeder2016basics}.

\xhdr{Lexicographic order}
Let $\seqa, \seqb$ with \mbox{$\len{\seqa} = \len{\seqb} = n$} be two sequences with elements $a_i, b_i \in (\sS, \leq)$.
We define the lexicographic order of $\seqa$ and $\seqb$ as~\cite[Def. 1.8]{Harzheim2005orders}
\begin{align}
\begin{split}
    \seqa > \seqb \Leftrightarrow& \exists \, i \in \nintegers: \eseqa_i > \eseqb_i \land \forall j < i: \eseqa_j = \eseqb_j\, , \\
    \seqa = \seqb \Leftrightarrow& \forall \, i \in \nintegers: \eseqa_i = \eseqb_i \, .
\end{split}
\end{align}
Intuitively, this defines a hierarchical order over the elements of every sequence by decreasing importance.

Let $(f_{fast}, f_{cheap}, f_{robust})$ be the sequence of lexicographic fitness values.
As the fourth robot is the fastest one and robot three is cheaper than robot two, while they are equally fast, we obtain the lexicographic order \mbox{$\robot_4 > \robot_3 > \robot_2 > \robot_1$}.
However, if we chose the lexicographic order to be $(f_{robust}, f_{fast}, f_{cheap})$, the ranking would be \mbox{$\robot_1 > \robot_4 > \robot_3 > \robot_2$}.

\xhdr{Pareto optimality}
Let $\seqf, \seqg$ with \mbox{$\len{\seqf} = \len{\seqg} = n$} be two sequences with \mbox{$\eseqf_i, \eseqg_i \in (\sF, \leq)$}.
We say ``$\seqf$ dominates $\seqg$'' if it contains a higher value for at least one objective and if all other objectives are at least equal.
Formally,
\begin{align}
    \seqf \cgt \seqg \Leftrightarrow \exists \, i \in \nintegers: \eseqf_i > \eseqg_i \land \forall \, i \in \nintegers: \eseqf_i \geq \eseqg_i \, .
\end{align}
An element $\seqf$ is Pareto-optimal with respect to a set $\sF$ if it is not dominated by any element in this set~\cite[eq. 4.59]{Boyd2023convex}.

Let $(f_{fast}, f_{cheap}, f_{robust})$ be a multiobjective fitness sequence for the exemplary robots.
We can see that $\robot_3 \cgt \robot_2$ since both robots are equally fast and robust, but robot three is cheaper.
However, robots one, three, and four are all Pareto-optimal as none is dominated by any other robot.

\xhdr{Genetic algorithm}
Genetic algorithms change the \textit{population} of different solution candidates (\textit{individuals}) to maximize an objective, called the \textit{fitness} function~\cite{Coello2000}.
Each individual is represented by a \textit{genome} that encodes its unique attributes.
For a fixed number of iterations, a new population is created by selecting a limited number of individuals from the previous population (selection), creating offsprings by combining attributes of existing individuals (crossover), and mutating some of the genomes (mutation).
A cycle of \textit{selection}, \textit{crossover}, and \textit{mutation} produces a new \textit{generation}.

\xhdr{NSGA-II}
We write $\ndr{\seqa, \sA}$ and $\crowdingdistance{\seqa, \sA}$ whenever we compute the nondomination rank or the crowding distance of an individual $\seqa$ with respect to a population $\sA$ according to the NSGA-II algorithm as proposed in~\cite{Deb2002}.
Appendix~\ref{appdx:nsga2_metrics} provides an informal introduction to these metrics.

\subsection{Lexicographic nondominant sorting optimization}~\label{sec:lex_opt_method}
\begin{table}
\caption{Notation}
\begin{tabularx}{\linewidth}{%
    >{\raggedright\arraybackslash}p{0.16\linewidth} %
    >{\raggedright\arraybackslash}p{0.29\linewidth} %
    >{\raggedright\arraybackslash}p{0.1\linewidth} %
    >{\raggedright\arraybackslash}X}
\toprule
\multicolumn{4}{l}{\textit{General notation}:} \\
    $\seqa, \seqb, \seqc$           & Sequences                     & $\sA, \sB, \sC$    & Sets \\
    $\va, \vb, \vc$                 & Vectors                       & $\mA, \mB, \mC$    & Matrices \\
    $\alpha, \beta, \gamma \in \R$  & Scalar variables              & $\msA, \msB, \msC$ & Multisets \\
    $\nintegers$                    & The set $\{1, 2, \dots, n\}$  & $(\sA, \leq)$      & Partially ordered set \\[0.5em]
\end{tabularx}
\begin{tabularx}{\linewidth}{%
    >{\raggedright\arraybackslash}p{0.17\linewidth} %
    >{\raggedright\arraybackslash}X}
\midrule
\multicolumn{2}{l}{\textit{Important symbols}:} \\
    $\vb$                   & Pose $(x, y, \theta)$ of the mobile base \\
    $\morphology$           & A sequence of modules, defining a robot \\
    $\moduleset$            & Available modules \\
    $\vq$                   & Joint angles of the manipulator \\
    $\jointspace$           & Robot joint space \\
    $\goal = (\pose, \tol)$ & A goal with desired pose $\pose$ and tolerance $\tol$ \\
    $\pose = (\destrans, \desrot)$ & A pose with translation $\destrans \in \R^3$ and orientation $\desrot \in SO(3)$ \\
    $\obstacles$            & Environment obstacles \\
    $\obstacleself{\vq}{i}$ & Space occupied by the $i$th link of $\morphology$ for joint angles $\vq$ \\[0.5em]
\midrule
\multicolumn{2}{l}{\textit{Custom Operators}:} \\
    $\crowdingdistance{\seqa, \sA}$ & The crowding distance of a multiobjective value $\seqa$ with respect to its nondomination front in set $\sA$ \\
    $\ndr{\seqa, \sA}$  & The nondomination rank of a multiobjective value $\seqa$ with respect to the set $\sA$ \\
    $\countmod{x}{\cdot}$ & The number of elements $x$ within a sequence ($\countmod{x}{\seqx}$) or multiset ($\countmod{x}{\msX}$) \\
    $\len{\seqa}$   & The number of elements in a sequence \\
    $\size{m}$      & The size of a robot module, i.e., the Euclidean distance between its proximal and distal connection interfaces \\
\bottomrule
\end{tabularx}

\label{tab:notation}
\end{table}
We perform a mixed Pareto-lexicographic optimization based on the method introduced in~\cite{Kuelz2024}, which presents a single-objective genetic algorithm with a purely lexicographic fitness function.
The introduction of a lexicographic fitness function is motivated by computational efficiency:
By introducing multiple surrogate objectives, the selection process of the genetic optimization can be performed without evaluating the computationally expensive primary objective for most solution candidates.

For every individual $i$ in a population, we compute two sequences $\seqc_i$ and $\seqx_i$ as follows:
The sequence $\seqc_i$ contains $m$ values that measure compliance with $m$ constraints.
Crucially, some of them are introduced because they can easily be evaluated, such as the sufficient size of a robot or the absence of self-collisions.
We define these constraints so that the $j$th value of $\seqc_i$, $c_{i, j}$ is $0$, if the solution $i$ satisfies the constraint $j$, and negative, otherwise.
Further, we introduce the sequence 
\begin{align}
    \seqx_i &= \begin{cases}
        (a_i, b_i, \dots)   &\text{, if } \seqc_{i} = \bm{0} \\
        (-\infty, -\infty, \dots)  &\text{, otherwise,}
    \end{cases} \label{eq:moo_objective_definition}
\end{align}
where $a_i, b_i, \dots$ are our optimization objectives, such as the monetary cost or robustness of an individual solution.
We write $\sX = \{\seqx_1, \dots, \seqx_n\}$ and define the lexicographic fitness as the sequence
\begin{align}
    \seqf_i = (
    \underbrace{\eseqc_{i, 1}, \eseqc_{i, 2}, \dots, \eseqc_{i, m}}_{\text{constraint satisfaction}},
    \underbrace{\ndr{\seqx_i, \sX}}_{\text{nondomination rank}},
    \underbrace{\crowdingdistance{\seqx_i, \sX}}_{\text{crowding distance}}) \, . \label{eq:fitnessfun}
\end{align}
By performing a lexicographic selection based on this fitness, we can efficiently rule out candidates that do not fulfill the constraints.
Further, the lexicographic selection process allows one to pick candidates that are closer to fulfilling the constraints if not enough feasible solutions exist.
In addition, the nondomniation rank fitness from ~\eqref{eq:fitnessfun} allows us to perform multiobjective selection considering $\seqx_1, \dots, \seqx_n$.
Instead of a single solution, our optimization results in a set of Pareto-optimal solutions from which an operator can pick.

\subsection{Task representation}\label{sec:task_repr}
We obtain a formalized task representation by parsing the BIM model and user input as described in Sec.~\ref{sec:framework}.
The representation of goals, obstacles, tolerances, and robots follows the modeling language introduced by the CoBRA benchmark~\cite{Mayer2024}.
A formal definition for our specific use case is included in Appendices \ref{appdx:cobra} and \ref{appdx:tolerances}.

\subsection{Fitness functions}\label{sec:fitness_funs}
We obtain base movement $\vb(t)$ and joint angles $\jointpath$ over time from a black-box path planner
\begin{align}
    \motionplan{\morphology, \goals, \constraints, t} = %
    \begin{bmatrix} \vb(t) \\ \jointpath \end{bmatrix} \, .
\end{align}
The planner guarantees satisfaction for all constraints $\constraints$ considered in this work, such as collision avoidance.
However, inverse kinematics solutions to reach all goals $\goals$ may not always exist.
Furthermore, since we rely on numeric optimization, we might not be able to find them with perfect accuracy.
To obtain a scalar expression of the errors $\poseerr{} = \left( \transerr, \roterr \right)$ defined in~\eqref{eqn:poseerr}, we introduce weighting factors $w_t, w_R$ and error measures $\projtrans: \R^3 \to \R$ and $\projrot: SO(3) \to \R$ and define the distance function
\begin{align}
    d(\poseerr) = w_t \projtrans(\transerr) + w_R \projrot(\roterr) \, .
\end{align}
While the measures can be chosen arbitrarily, we usually use the Euclidean norm for $\projtrans$ and define $\projrot$ as the Euler angle of rotation, as detailed in~\eqref{eq:axis_angle}.
The end-effector poses can be easily computed using forward kinematics $FK: \R^3 \times \jointspace \to SE(3)$.
This allows us to obtain the minimum pose error for goal $\goal_i$:
\begin{align}
    \poseerr^*_i &= \poseerr{\Bigl(\pose_i, FK \bigl(\motionplan{\hat{t}} \bigr) \Bigr)} \text{, where} \\
    \hat{t} &= \argmin_t d \Bigl( \poseerr \bigl( \pose_i, FK(\motionplan{t}) \bigr) \Bigr) \, \nonumber .
\end{align}
If the goal error is within the goal tolerance for every goal in a task, we say that the path $\path$ is feasible.

Let $\obstacleself{\vq}{i}$ be the space occupied by the $i$th link of robot $\morphology$ with joint configuration $\vq$.
The following constraints for
joint limits~\oldeqref{eq:clim},
self-collisions~\oldeqref{eq:sc_constraint},
collisions with the environment~\oldeqref{eq:col_constraint},
and torque limits~\oldeqref{eq:torque_constraint},
are later used for the constraint functions $c_j$ and all evaluate to true or false:
\begin{alignat}{3}
    & c_{\text{lim}} && \iff \forall t \in [T]: \jointpath \in \jointspace \label{eq:clim} \\
    & c_{\text{sc}}  && \iff \forall \left(i, t\right) \in \nintegers \times [T], \, \forall j < i: \nonumber \\
    & && \qquad \qquad \obstacleself{\jointpath}{i} \cap \obstacleself{\jointpath}{j} = \emptyset  \label{eq:sc_constraint} \\
    & c_{\text{col}} && \iff \forall \left(i, t\right) \in \nintegers \times [T]: \obstacles \cap \obstacleself{\path}{i} = \emptyset \label{eq:col_constraint} \\
    & c_{\tau} && \iff \forall t \in [T]: \boldsymbol{\tau}(t, \vf_{ext}(t)) \in [\underline{\boldsymbol{\tau}}, \overline{\boldsymbol{\tau}}] \label{eq:torque_constraint}
    \, .
\end{alignat}
Here, $\boldsymbol{\tau}(t, \bm{f}_{ext}(t))$ denotes robot joint torques under consideration of an external payload $\vf_{ext}(t)$.
We define the fitness functions for the constraint satisfaction of the robot $\morphology$ as follows.
\begin{align}
    {c_1} &= \min \left( 0, \sum_{i=1}^{\len{\morphology}} \bigg( \size{\eseqm_i} \bigg) - \max_{i \in [|\goals|]} \left( \norm{\destrans_i}_2 \right) \right) \label{eq:c_size} \\
    {c_2} &= \min \left( 0, \min_{i} \left( \countmod{\eseqm_i}{\moduleset} - \countmod{\eseqm_i}{\morphology} \right) \right) \label{eq:c_modules} \\
    {c_3} &= -\sum_{t > T_{cal}} \norm{\vb(t) - \vb(T_{cal})} \label{eq:c_basemove} \\
    {c_4} &= \begin{cases}
        0 &\text{, if } \feasible{\motionplan{\morphology, \goals, \{c_{\text{lim}}\}, \goals, t}} \\
        -d(\poseerr^*_{i}) &\text{, otherwise.}
    \end{cases} \label{eq:c_jointlim} \\
    {c_5} &= \begin{cases}
        0 &\text{, if } \feasible{\motionplan{\morphology, \goals, \{c_{\text{lim}}, c_{\text{sc}}\}, \goals}, t} \\
        -d(\poseerr^*_{i}) &\text{, otherwise.}
    \end{cases} \label{eq:c_selfcol} \\
    {c_6} &= \begin{cases}
        0 &\text{, if } \feasible{\motionplan{\morphology, \goals, \{c_{\text{lim}}, c_{\text{sc}}, c_{\text{col}}\}, \goals, t}} \\
        -d(\poseerr^*_{i}) &\text{, otherwise.}
    \end{cases} \label{eq:c_envcol} \\
    c_7 & 
    =
    \sum_{t=1}^T
    \sum_{i=1}^n
    \min \left(
        0,
        \overline{\tau}_i - \tau_i(t, \vf_{ext}),
        \tau_i(t, \vf_{ext}) - \underline{\tau}_i
    \right)
    \label{eq:c_torque}
\end{align}
The functions in~\plaineqref{eq:c_size} and~\plaineqref{eq:c_modules} can be evaluated without obtaining a kinematic or dynamic model of the robot and indicate whether the individual modules of the robot are large enough to reach all goals and if they are available on-site, respectively.
The fitness function in~\plaineqref{eq:c_basemove} forces the base not to move after visual calibration is performed ($t=T_{cal}$, see also~\figref{fig:drilling_bt}).
The next three functions successively assess whether the joint path is valid with respect to joint limits~\plaineqref{eq:c_jointlim}, self-collisions~\plaineqref{eq:c_selfcol}, and collisions with the environment~\plaineqref{eq:c_envcol}.
Finally, \plaineqref{eq:c_torque} is zero if and only if all torque limits are kept.

If all constraints are satisfied, we compute the values for our multiobjective fitness criteria.
The compactness
\begin{align}
    f_{c}  &= -\sum_{i=1}^n \size{m_i}
\end{align}
indicates the negative value of the total size of all modules combined.
High compactness usually allows for faster transport and a shorter lever for the forces induced by drilling.

The robustness fitness function indicates the maximum placement error \mbox{$\Delta_b = (\Delta_x, \Delta_y, \Delta_\theta)^T$} our motion planner can account for online.
The adjustment function \mbox{$\mathrm{replan}(\path, \Delta_b)$} indicates whether there is an adjustment that when applied to the joint path $\jointpath$, results in the end effector of the robot still following the desired motion.
Finally, we define the robustness of a robot $\morphology$ with respect to positioning errors of the mobile base as
\begin{align}
    f_{r}   = \max \{ &\delta \in [0, 1] : \label{eqn:robustness_score} \\
    & \quad \mathrm{replan}(\motionplan{\morphology, \goals, \constraints, t}, \delta \Delta_{max}) = \mathtt{true} \} \, . \nonumber
\end{align}
The maximum displacement $\Delta_{max}$ can be chosen based on the available hardware and prior experience.
We chose $\begin{bmatrix}20\text{cm} & 20\text{cm} & 15\deg\end{bmatrix}^T$ for our experiments.
The above definition does not discriminate between robustness against positioning errors in the $xy$-plane and orientation errors $\Delta_\theta$.
If necessary, the robustness score can be easily extended to multiple scores, one for every possible error.

Lastly, the reconfiguration time fitness indicates the time necessary to reconfigure the current module composition $\morphology$ to match a new desired configuration $\morphology_d$.
If the current composition is unknown or nonexistent, for example, because the modules are currently disassembled, we set $\morphology$ to the empty tuple, so the reconfiguration fitness measures the initial time for setup.
Assume that $\morphology$ and $\morphology_d$ share $n_c$ common modules at the beginning of their kinematic chain.
Then, the reconfiguration time fitness is chosen as
\begin{align}
    f_{t} &= -t_m \left( \len{\morphology} + \len{\morphology_d} - 2 n_c \right) \, .
\end{align}
Here, we assume that removing or adding a module takes time $t_m$, regardless of the module.

\section{Experiments}\label{sec:experiments}
We conducted experiments on a drilling use case in the Fraunhofer Italia ARENA at the NOI Techpark in Bolzano, Italy.
The experimental setup is shown in \figref{fig:expSetup}.
It consists of a movable wall with four blocks (brick, sandstone brick, wood, and insulation material) that is modeled in the used \ac{bim} file\footnote{
\url{https://github.com/FraunhoferItalia/rosbim/tree/humble-dev/rosbim_example_models}
}.

\subsection{Robot modules}\label{sec:robot_modules}
We use the robot modules introduced in~\cite{Romiti2022} and shown in Figure~\ref{fig:expSetup}A, together with an RB-VOGUI mobile base\footnote{
\url{https://robotnik.eu/products/mobile-robots/rb-vogui-en/}
} with a custom cabinet top setup and two lidars (SICK expert S300).
In total, we use one mounting base, three elbow- and straight joint modules each, three passive links\footnote{
\url{ https://github.com/FraunhoferItalia/fhi_resources}
}
with a length of $100$, $200$, and $500$ millimeters, respectively, as well as a custom-made end effector composed of a drill tool (Einhell Herocco 18/20 with custom 3D printed support) and an RGB-D camera (Intel Realsense D435i).

\subsection{Drilling in simulation}\label{sec:drilling_simulation}
\begin{figure}
    \centering
    
    \begin{minipage}{\linewidth}
    \centering
    \subfloat[%
        Normalized compactness and robustness of Pareto-optimal results within 100 individual runs.
        The heights of the boxes indicate the relative frequency of a particular fitness score observed.
        The green line indicates the Pareto optimal results for all trials.
    \label{fig:num_exp}]{
        \includegraphics[width=0.94\linewidth]{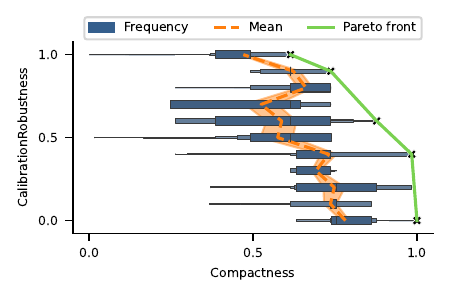}
    }
    \end{minipage}%
    \hfill%
    \begin{minipage}{\figwdefault}
    \centering
    \subfloat[%
        The best robustness score that could be achieved with a limited reconfiguration time.
    \label{fig:reconfig_robustness}]{
        \includegraphics[width=0.94\linewidth]{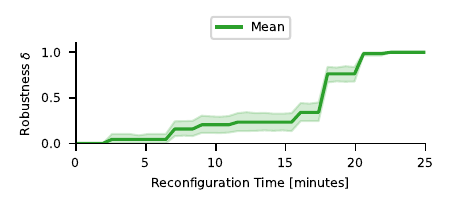}
    }
    \end{minipage}
    \caption{%
        Aggregated results for 100 different random initial populations.
        Shaded areas indicate 95\% confidence intervals using bootstrapping.
    }
\end{figure}
In simulation, we optimized robots to drill two holes of 15\,cm depth in the sandstone brick.
Based on the maximum measurements in preliminary experiments, we model the external payload during drilling as a constant external force of $13\text{N}$ and torque of $15\text{Nm}$ acting along and about the drill axis, respectively.
We assume no external payload for other parts of the trajectory.
To assess the robustness of a solution against base positioning errors, we changed the base position after calibration by a factor of $\delta\,20\,$cm in both horizontal directions and rotated it by $\delta \,15^\circ$, where $\delta \in [0, 1]$ defines the desired level of robustness according to~\eqref{eqn:robustness_score}.
Details about the tolerances used in the simulation are listed in Appendix~\ref{appdx:tolerances}.
To analyze the relationship between the different objectives, we conducted 100 optimization trials, each starting with a different random initialization.
We limited computational resources per trial to $50$ generations, resulting in a mean runtime of $78$ minutes.

Each of the 100 runs resulted in at least eleven valid solutions, all of which had five or six joints.
On average, the Pareto front consisted of $3.7$ distinct individuals.
For $99\%$ of the Pareto-optimal solutions, the first two joints were a straight, then an elbow joint.
Performance differences resulted from the remaining modules, which were distributed in various ways.
Interestingly, we observed a strong positive correlation between solution robustness and using the longest passive link (Pearson correlation coefficient $\rho = 0.54, p < 0.01$).
The robustness of the calibration was also correlated with the use of the third straight joint ($\rho = 0.32, p<0.01$).
In contrast, using the third elbow joint was positively correlated with the compactness of Pareto-optimal solutions ($\rho = 0.52, p<0.01$).
\Figref{fig:num_exp} shows the Pareto-optimal solutions of each trial regarding compactness and robustness.
The compactness is normalized for better readability; that is, $0$ indicates the largest robot and $1$ the most compact robot present in the final solution set, respectively.
As our optimization is not complete, the theoretic Pareto front remains unknown.
Instead, we report the evidence-based Pareto front by aggregating all trials.
Further, \figref{fig:reconfig_robustness} shows the trade-off between robustness and reconfiguration time, starting from an exemplary initial configuration.

In addition to our algorithm, we performed ten optimizations using the lexicographic genetic approach introduced in~\cite{Kuelz2024}.
As this algorithm does not support multiple objectives, we optimized for compactness only.
Other fitness functions, constraints, and hyperparameters remained unchanged.
Normalized by the results of our algorithm, we observed an average compactness score of $0.85$ (maximum: $1.0$) and an average robustness score of $0.04$ (maximum: $0.1$).

\subsection{Real-world experiments}
\begin{figure*}
    \centering
    \includegraphics[width=.9\linewidth,trim=40 100 40 30,clip]{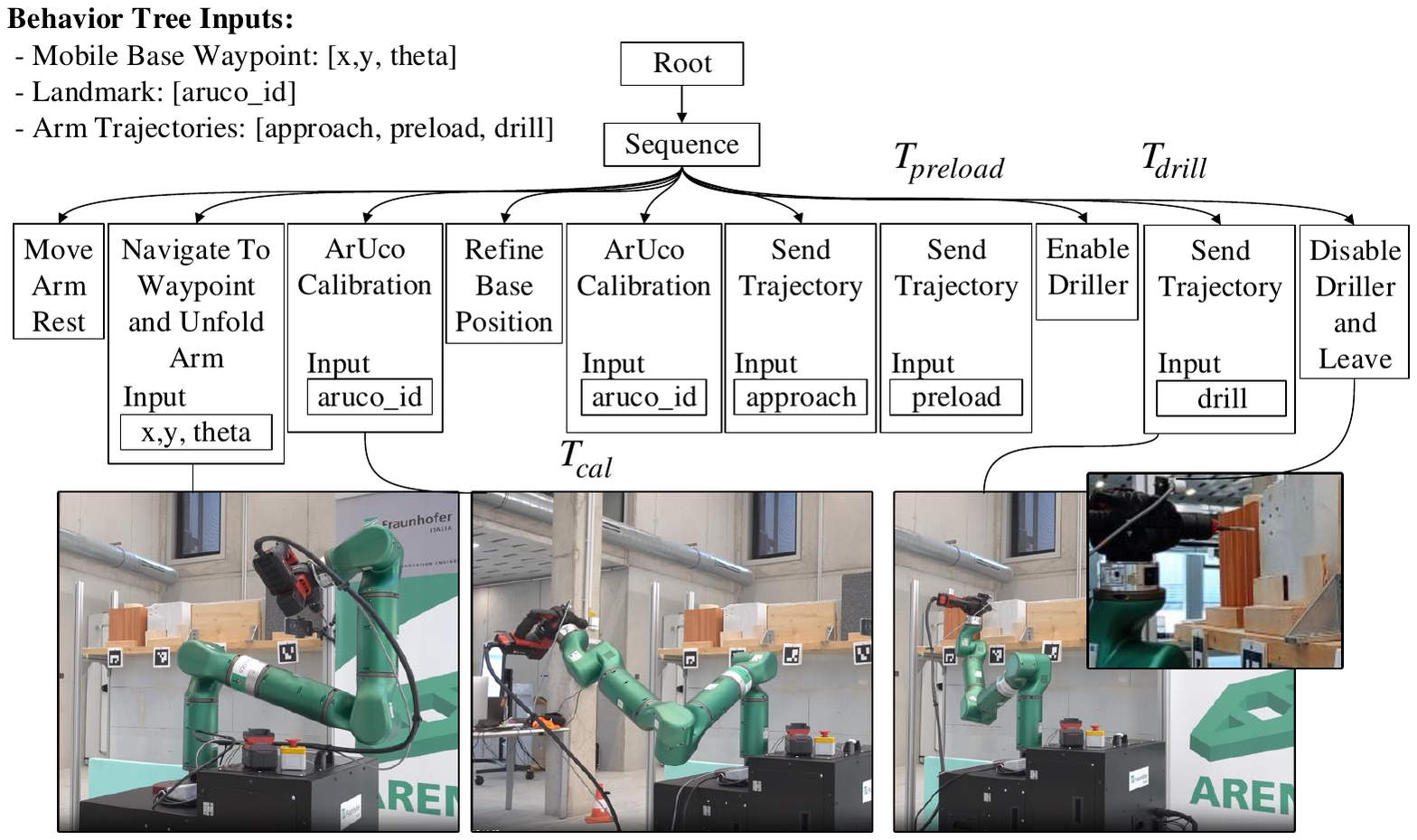}
    \caption{The simplified behavior tree of the drilling experiment on a sandstone brick with the six-degrees-of-freedom configuration of the modular robotic arm.}
    \label{fig:drilling_bt}
\end{figure*}

\begin{figure}
    \centering
    \includegraphics[width=0.75\figwdefault]{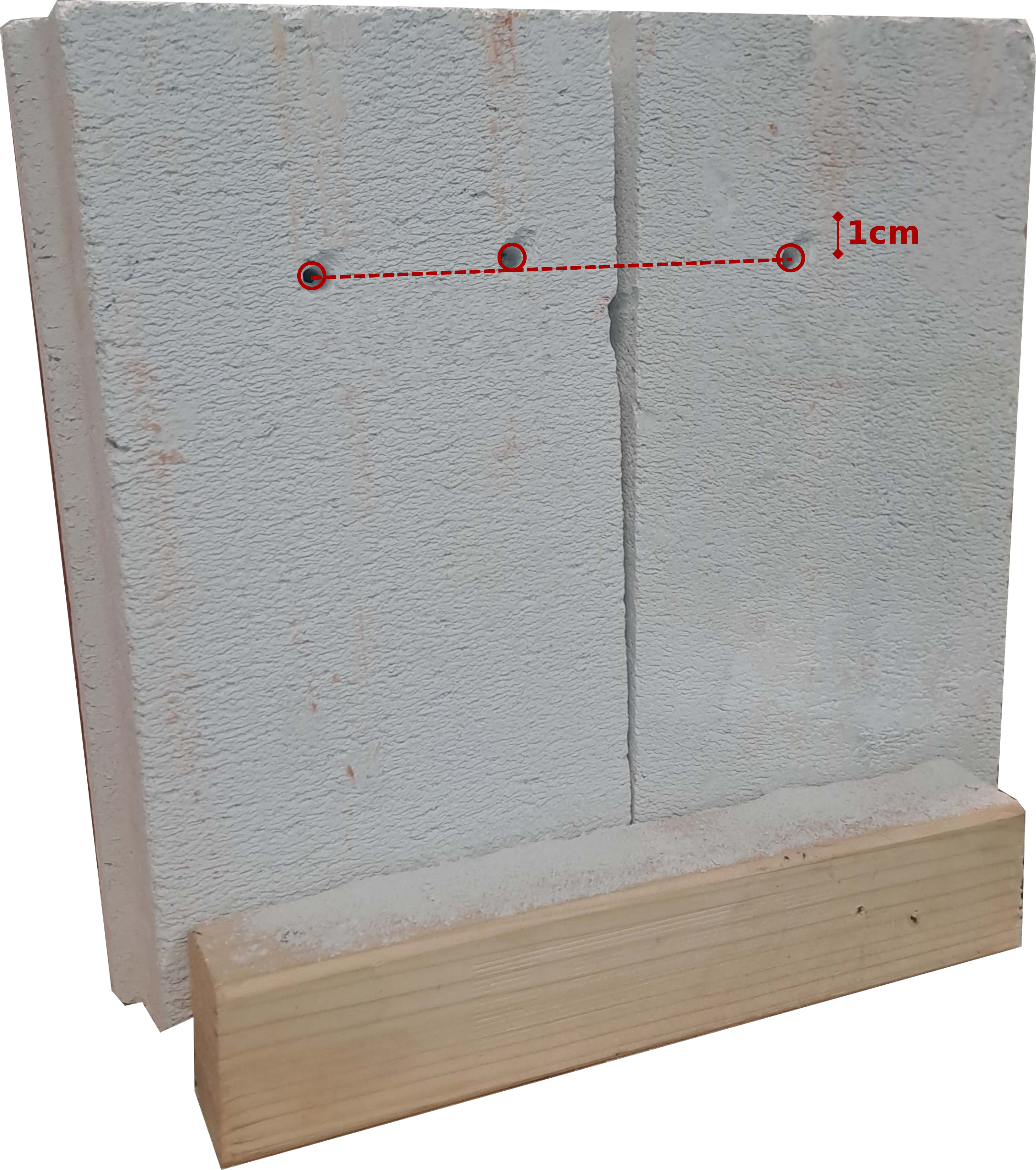}
    \caption{%
    The sandstone bricks after drilling three holes in a vertical line.
    The repeatability precision of the whole approach, including navigation and calibration is within $1$\,cm.
    }
    \label{fig:precision}
\end{figure}

To evaluate the real-world performance of our system, we conducted a drilling experiment covering the entire workflow from task specification to execution.
\Figref{fig:drilling_bt} shows a simplified version of the behavior tree that consists of one sequence of robot actions used in the experiments. The full tree also includes fallback nodes for error handling.
The drilling task includes multiple stages.
First, the mobile robot base navigates towards the experimental setup, using waypoints identified from the BIM model.
The robot arm then unfolds to a calibration pose, and the base position is refined based on the ArUco markers, scanned by a camera mounted on the end effector.
After calibration, the initial planned robot trajectory is adapted online to counteract the final positioning error of the mobile base.
The robustness of the robot plays a crucial role in this step.
If the initial trajectory is close to the joint limits of the robot, a large positioning error might lead to failure at this stage, requiring recalibration.
Finally, the robot performs the approach and drill motions, where the trajectory is executed using the impedance controller of the robot arm.
Pictures of the drilled holes can be seen in the most-right snapshot in \Figref{fig:drilling_bt} and in \figref{fig:precision}.
A complete video of the mission execution is available on the project website \mbox{\href{https://modular-construction-robot.cps.cit.tum.de}{modular-construction-robot.cps.cit.tum.de}}.
\begin{figure}
    \centering
    \includegraphics[width=\sixthfigurewidth]{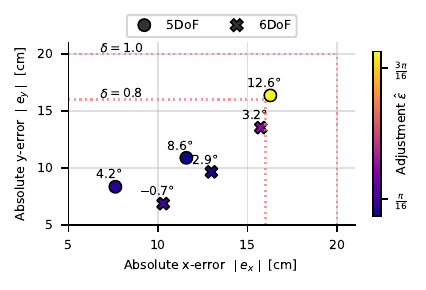}
    \caption{%
    Among the conducted experiments, we measure absolute errors between $6$\,cm and $16$\,cm regarding the base position and between $0.7^\circ$ and $12.6^\circ$ regarding its orientation.
    The figure shows the positioning errors of the base with respect to the ArUco markers and the maximum joint angle adjustment during the execution of the drilling task.
    }
    \label{fig:arm_correction}
\end{figure}

Based on previous experience with the mobile base, we chose a minimum robustness score of $\delta = 0.8$.
Consecutively, we selected two solutions from our optimization: one with a six-degree-of-freedom (DoF) robot arm and one with a five-DoF robot arm.
For each configuration, we performed three drilling missions on the sandstone brick.

\Figref{fig:arm_correction} shows the absolute error of the mobile base position relative to the nominal drilling position for each experiment, expressed in terms of $e_x$, $e_y$, and $\theta$.
Moreover, it indicates the adjustment necessary to correct the initial motion plan according to the positioning error.
We compute this score as the maximum change in a single joint angle over all joint configurations in the drilling path $\path$, i.e.,
\begin{align}
    \hat{\varepsilon} = \max_{t \in [T]} \left( \norm{\vq(t) - \hat{\vq}(t)}_\infty \right) \, ,
\end{align}
where $\vq$ is the initially planned joint configuration, and $\hat{\vq}$ is the one observed during execution.
We chose this metric as it can easily be compared between robots with varying DoF; however, the general trend shown in \figref{fig:arm_correction} persists over different choices of norms.

\Figref{fig:precision} shows the sandstone brick after drilling three holes in a horizontal line.
The relative vertical precision was within one centimeter.
Initially, the modules were disassembled, as depicted in Figure~\ref{fig:expSetup}A.
Upon receiving the configuration for the first robot (a six-DoF setup consisting of the following module sequence: straight, elbow, passive 100mm, straight, elbow, straight, elbow, end effector), the operator assembled the robot in ten minutes.
The second robot consisted of the following module sequence: straight, elbow, passive 100mm, straight, elbow, elbow, end effector.
The time required to reconfigure the robot and launch the mission was reduced to about five minutes, since the first five modules were shared with the previous configuration.

\subsection{Spray painting in simulation}
\begin{figure}
    \centering
    \includegraphics[width=0.65\linewidth]{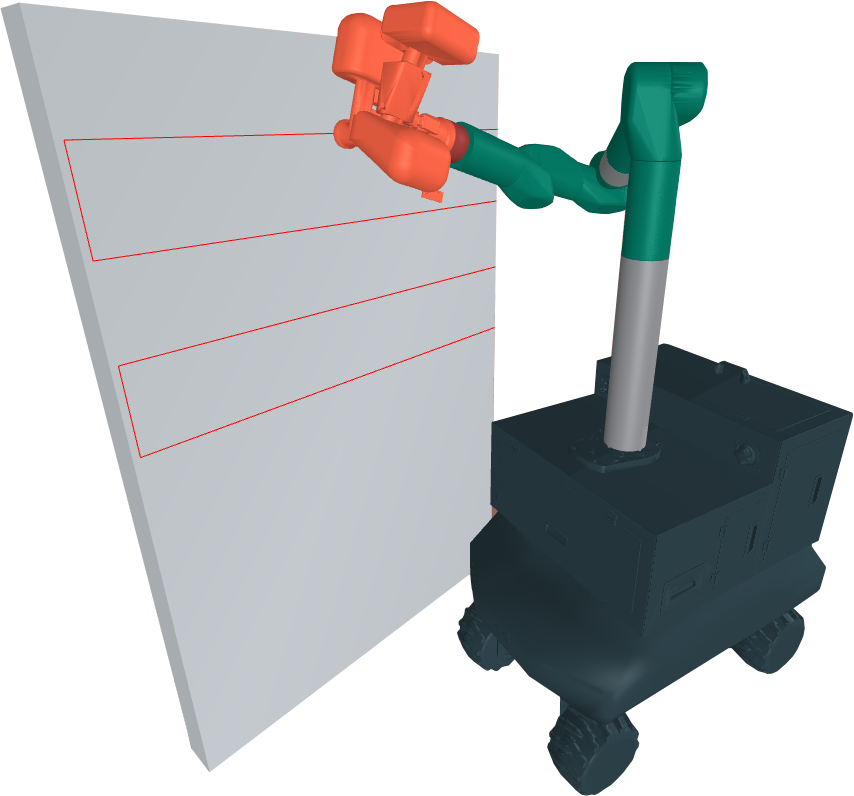}
    \caption{Simulation of spray painting:
    The thin red line shows the target trajectory.}
    \label{fig:spraypaint_simulation}
\end{figure}
\begin{figure}
    \centering
    \includegraphics[width=.94\linewidth]{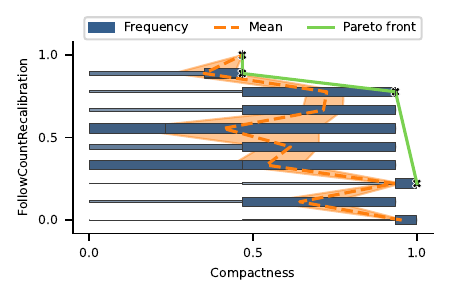}
    \caption{Normalized compactness and robustness of Pareto-optimal results for spray painting within 100 individual runs.
    The heights of the boxes indicate the relative frequency of a particular fitness score observed.
    The green line indicates the Pareto optimal results for all trials.}
    \label{fig:numerical_results_painting}
\end{figure}
Our framework can be adapted to various construction tasks, as shown by the following example for spray painting.
We adapt the BIM model of the drilling task by including an additional wall.
Additionally, we replaced the end-effector module with a custom end effector for spray painting, first introduced in~\cite{Terzer2024}.
In a slightly modified user interface, users can select a rectangular area to be painted instead of drilling positions, for which a trajectory covering this area is automatically computed.
We also adjust the constraints by removing the drilling payload and allowing the robot base to move during task execution to cover larger areas.
Similarly to drilling, task performance is independent of the end-effector orientation around its local z-axis, so the goal tolerances need not be adjusted.
Other than in the drilling task, the main challenge in spray painting is not the accurate positioning of the end effector.
Instead, reducing the number of recalibrations triggered by base movement is important.
Consequently, we use the negative number of required recalibrations as the robustness score for this task.

We performed 100 optimization trials, 99 of which resulted in at least one valid robot morphology.
On average, the Pareto front consisted of 3.1 distinct individuals.
In contrast to the drilling task, 18\% of the robots had a passive link as their first module, as shown in Figure~\ref{fig:spraypaint_simulation}.
The robots for spray painting consisted of an average of $8.2$ modules, compared to an average of $6.5$ for drilling.
Figure~\ref{fig:numerical_results_painting} shows the normalized scores for Pareto-optimal solutions for spray painting.
Similarly to the drilling task, we observed a trade-off between the compactness and robustness of the solutions.
Again, this was also reflected in a positive correlation between the use of the longest passive link and the robustness $(\rho = 0.41, p < 0.01)$.
All valid solutions used the three elbow joints available.

\subsection{Discussion}
The results show the trade-offs between compactness, robustness, and configuration time.
As shown by the results for our baseline algorithm, single-objective optimization approaches tend to provide a single solution on the Pareto front only.
Theoretically, the full Pareto front can still be recovered using single-objective optimization.
However, this would require carefully chosen weighting factors and scale exponentially with the number of objectives to be considered.
Our results show that explicit knowledge about the trade-off between competing objectives is crucial for practical applications.
The Pareto front shown in Figure~\ref{fig:num_exp} indicates that a good robustness score of $0.4$ can already be achieved with a robot almost as compact as the most compact candidate.
This relation does not hold the other way around:
The most robust robots are significantly less compact than those with robustness scores of $0.9$ or less.

When starting from an assembled robot from a previous task, the mechanical assembly of the first valid robot takes three minutes.
However, if high robustness is essential, a reconfiguration time of more than 20 minutes is necessary to assemble the corresponding robot\footnote{Note that in general, setting up a new modular robot takes around half that time. In this experiment, the most robust solutions required complete disassembly of an existing configuration, resulting in a long reconfiguration time.}.
Both results underline the benefits of multiobjective optimization:
Based on the expected disturbances, the required compactness, and the time at hand, an operator can choose the situationally optimal solution for deployment.

The results of our simulations show the clear benefit of algorithmic support in this process, as some of the outcomes contradict initial human intuition.
When adapting a robot to be more robust against calibration errors, the first reflex would often consist of adding more joints.
As indicated by the observed trade-off between compactness and robustness, a correlation exists between more modules and higher robustness.
However, our experiments have shown that the additional mass and lever of a new module often lead to a violation of the joint torque constraint.
Further, the changed kinematics usually resulted in a complete recomputation of the motion plan, which is often less robust due to self-collisions.
Our results show that the use of long passive links often leads to high robustness, as they increase the reach of the robot while adding only a fraction of the mass of a new joint.
Thus, to achieve a more robust robot, a complete reconfiguration might be necessary, as further highlighted in Figure~\ref{fig:reconfig_robustness}.
Design patterns, such as a minimum required number of joints or an almost consistent choice for the first two joints, emerged, indicating the existence of specific dominant configurations.
However, as demonstrated by the spray painting simulation, these patterns are task-specific.

The robot could adjust for mobile base positioning errors in all real-world experiments.
As expected, the adjustment increases with higher positioning errors.
Furthermore, as shown in \figref{fig:arm_correction}, the absolute errors differ between runs, and for five of the six experiments, the necessary adjustment was below 50\% of the maximum adjustment.
If a more compact (and thereby less robust) robot along the Pareto front needs to be chosen, high positioning errors of the base could likely be overcome by reapproaching the wall and repeating the calibration.

The primary purpose of the experiments conducted was to demonstrate the feasibility of the proposed framework.
The control gains were intentionally set to a lower level to prevent excessive force, resulting in reduced precision.
This ensures that the robot does not push too far into the wall, even if the vision system fails to accurately detect the marker.
Still, as shown in \figref{fig:precision}, we observed a relative horizontal precision of $1$\,cm.
In future applications, the system could be enhanced with a more accurate vision system to improve its accuracy.
Furthermore, implementing a force control mechanism would further robustify the system and enable a more precise and adaptive interaction with the environment.

\section{Conclusion and Limitations}\label{sec:conclusion}
We developed and integrated a holistic framework that allows construction workers to instruct a mobile modular robot arm for building construction.
In contrast to existing solutions, the modularity of hardware and software components and multiobjective optimization of robot morphology enables tailoring the proposed approach to specific needs in situ.
By integrating \ac{bim} information into optimization and control frameworks from the robotics domain, we have created a holistic framework that requires minimal intervention and robotics expertise for deployment.
In addition, our approach is not limited to considering geometric data of the used BIM models; future work could also incorporate information about material properties to enable automation even for dynamically more challenging tasks, such as impact drilling.

Evolutionary approaches do not provide guarantees regarding their optimality or completeness.
In addition, the quality of the solutions depends not only on our method but also on the underlying motion planner used for trajectory generation.
Lastly, a successful transfer to the real world depends on the quality of the BIM data and the sensors.
While the overall approach applies to various tasks, future applications, such as bricklaying or plastering, will require adjusted simulation models, constraints, and optimization objectives.

\section*{Acknowledgment}
The authors gratefully acknowledge financial support by the Horizon 2020 EU Framework Project CONCERT under grant 101016007 and by the Deutsche Forschungsgemeinschaft (German Research Foundation) under grant number AL 1185/31-1.
We also thank Matthias Mayer for valuable feedback and discussions and Julius Emig for the help on the experimental setup.

\bibliographystyle{IEEEtran}
\bibliography{bibliography}

\begin{thebibliography}{10}
\providecommand{\url}[1]{#1}
\csname url@rmstyle\endcsname
\providecommand{\newblock}{\relax}
\providecommand{\bibinfo}[2]{#2}
\providecommand\BIBentrySTDinterwordspacing{\spaceskip=0pt\relax}
\providecommand\BIBentryALTinterwordstretchfactor{4}
\providecommand\BIBentryALTinterwordspacing{\spaceskip=\fontdimen2\font plus
\BIBentryALTinterwordstretchfactor\fontdimen3\font minus \fontdimen4\font\relax}
\providecommand\BIBforeignlanguage[2]{{%
\expandafter\ifx\csname l@#1\endcsname\relax
\typeout{** WARNING: IEEEtran.bst: No hyphenation pattern has been}%
\typeout{** loaded for the language `#1'. Using the pattern for}%
\typeout{** the default language instead.}%
\else
\language=\csname l@#1\endcsname
\fi
#2}}

\bibitem{Graetz2018}
G.~Graetz and G.~Michaels, ``Robots at work,'' \emph{The Review of Economics and Statistics}, vol. 100, no.~5, pp. 753--768, 2018.

\bibitem{Kim2015}
M.~J. Kim, \emph{et~al.}, ``Automation and robotics in construction and civil engineering,'' \emph{Journal of Intelligent \& Robotic Systems}, vol.~79, no.~3, pp. 347--350, 2015.

\bibitem{Delgado2019}
J.~M.~D. Delgado, \emph{et~al.}, ``Robotics and automated systems in construction: Understanding industry-specific challenges for adoption,'' \emph{Journal of Building Engineering}, vol.~26, 2019, article no. 100868.

\bibitem{AutomationHandbookConstructionAutomation}
D.~Castro-Lacouture, ``Construction automation and smart buildings,'' in \emph{Springer Handbook of Automation}, 2nd~ed., S.~Y. Nof, Ed., 2023, pp. 1035--1053.

\bibitem{Gappmaier2024}
P.~Gappmaier, S.~Reichenbach, and B.~Kromoser, ``Advances in formwork automation, structure and materials in concrete construction,'' \emph{Automation in Construction}, vol. 162, 2024, article no. 105373.

\bibitem{Bock2015}
T.~Bock and T.~Linner, \emph{Robot-Oriented Design}, T.~Bock, Ed.\hskip 1em plus 0.5em minus 0.4em\relax Cambridge University Press, 2015, vol.~1.

\bibitem{Ardiny2015}
H.~Ardiny, S.~Witwicki, and F.~Mondada, ``Construction automation with autonomous mobile robots: A review,'' in \emph{Proc. of the RSI Int. Conf. on Robotics and Mechatronics (ICROM)}, 2015, pp. 418--424.

\bibitem{Slongo2024}
C.~Slongo, \emph{et~al.}, ``Integrating automation into construction site: A system approach for the brick cutting use case,'' in \emph{Proc. of the Int. Symp. on Automation and Robotics in Construction (ISARC)}, 2024, pp. 251--258.

\bibitem{Gawel2019}
A.~Gawel, \emph{et~al.}, ``A fully-integrated sensing and control system for high-accuracy mobile robotic building construction,'' in \emph{Proc. of the IEEE/RSJ Int. Conf. on Intelligent Robots and Systems (IROS)}, 2019, pp. 2300--2307.

\bibitem{Doerfler2016}
K.~D\"{o}rfler, \emph{et~al.}, ``Mobile robotic brickwork,'' in \emph{Robotic Fabrication in Architecture, Art and Design}, D.~Reinhardt, R.~Saunders, and J.~Burry, Eds., 2016, pp. 204--217.

\bibitem{Liao2023}
H.~Liao, \emph{et~al.}, ``Robot-assisted after-process progress-monitoring system based on {BIM} and computer vision,'' in \emph{Proc. of the IEEE Int. Conf. on Robotics and Biomimetics (ROBIO)}, 2023, pp. 1--6.

\bibitem{Poirier2015}
E.~A. Poirier, S.~Staub-French, and D.~Forgues, ``Measuring the impact of {BIM} on labor productivity in a small specialty contracting enterprise through action-research,'' \emph{Automation in Construction}, vol.~58, pp. 74--84, 2015.

\bibitem{Giusti2021}
A.~Giusti, \emph{et~al.}, ``{BALTO}: A {BIM}-integrated mobile robot manipulator for precise and autonomous disinfection in buildings against {COVID}-19,'' in \emph{IEEE Int. Conf. on Automation Science and Engineering (CASE)}, 2021, pp. 1730--1737.

\bibitem{Chen2018}
Q.~Chen, B.~Garc\'{i}a~de Soto, and B.~T. Adey, ``Construction automation: Research areas, industry concerns and suggestions for advancement,'' \emph{Automation in Construction}, vol.~94, pp. 22--38, 2018.

\bibitem{Terzer2024}
M.~Terzer, \emph{et~al.}, ``A facilitated construction robot programming approach using building information modelling,'' in \emph{Proc. of the Int. Conf. on Control, Decision and Information Technologies (CoDIT)}, 2024, pp. 2656--2661.

\bibitem{Dautzenberg2023}
R.~Dautzenberg, \emph{et~al.}, ``A perching and tilting aerial robot for precise and versatile power tool work on vertical walls,'' in \emph{Proc. of the IEEE/RSJ Int. Conf. on Intelligent Robots and Systems (IROS)}, 2023, pp. 1094--1101.

\bibitem{Ortner2023}
M.~Ortner and B.~Kromoser, ``Influence of different parameters on drilling forces in automated drilling of concrete with industrial robots,'' \emph{Automation in Construction}, vol. 150, 2023.

\bibitem{Helm2012}
V.~Helm, \emph{et~al.}, ``Mobile robotic fabrication on construction sites: {dimRob},'' in \emph{Proc. of the IEEE/RSJ Int. Conf. on Intelligent Robots and Systems (IROS)}, 2012, pp. 4335--4341.

\bibitem{Outon2019}
J.~Outón, \emph{et~al.}, ``Innovative mobile manipulator solution for modern flexible manufacturing processes,'' \emph{Sensors}, vol.~19, no.~24, 2019, article no. 5414.

\bibitem{Stibinger2021}
P.~Štibinger, \emph{et~al.}, ``Mobile manipulator for autonomous localization, grasping and precise placement of construction material in a semi-structured environment,'' \emph{IEEE Robotics and Automation Letters}, vol.~6, no.~2, pp. 2595--2602, 2021.

\bibitem{Pankert2022}
J.~Pankert, \emph{et~al.}, ``Design and motion planning for a reconfigurable robotic base,'' \emph{IEEE Robotics and Automation Letters}, vol.~7, no.~4, pp. 9012--9019, 2022.

\bibitem{Buchli2018}
J.~Buchli, \emph{et~al.}, ``Digital in situ fabrication - challenges and opportunities for robotic in situ fabrication in architecture, construction, and beyond,'' \emph{Cement and Concrete Research}, vol. 112, pp. 66--75, 2018.

\bibitem{Paredis1993}
C.~J.~J. Paredis and P.~K. Khosla, ``Synthesis methodology for task based reconfiguration of modular manipulator systems,'' in \emph{Proc. of the Int. Symp. on Robotics Research (ISRR)}, 1993.

\bibitem{Han1997}
J.~Han, \emph{et~al.}, ``Task based design of modular robot manipulator using efficient genetic algorithm,'' in \emph{Proc. of the IEEE Int. Conf. on Robotics and Automation (ICRA)}, 1997, pp. 507--512.

\bibitem{Lipson2000}
H.~Lipson and J.~B. Pollack, ``Automatic design and manufacture of robotic lifeforms,'' \emph{Nature}, vol. 406, pp. 974--978, 2000.

\bibitem{Yim2007}
M.~Yim, \emph{et~al.}, ``Modular self-reconfigurable robot systems [grand challenges of robotics],'' \emph{IEEE Robotics and Automation Magazine}, vol.~14, no.~1, pp. 43--52, 2007.

\bibitem{Althoff2019}
M.~Althoff, \emph{et~al.}, ``Effortless creation of safe robots from modules through self-programming and self-verification,'' \emph{Science Robotics}, vol.~4, no.~31, 2019, article no. eaaw1924.

\bibitem{Chen1998}
I.-M. Chen and J.~W. Burdick, ``Enumerating the non-isomorphic assembly configurations of modular robotic systems,'' \emph{The International Journal of Robotics Research}, vol.~17, no.~7, pp. 702--719, 1998.

\bibitem{Icer2016}
E.~Icer and M.~Althoff, ``Cost-optimal composition synthesis for modular robots,'' in \emph{Proc. of the IEEE Conf. on Control Applications (CCA)}, 2016, pp. 1408--1413.

\bibitem{Liu2020}
S.~B. Liu and M.~Althoff, ``Optimizing performance in automation through modular robots,'' in \emph{Proc. of the IEEE Int. Conf. on Robotics and Automation (ICRA)}, 2020, pp. 4044--4050.

\bibitem{Icer2017}
E.~Icer, \emph{et~al.}, ``Evolutionary cost-optimal composition synthesis of modular robots considering a given task,'' in \emph{Proc. of the IEEE/RSJ Int. Conf. on Intelligent Robots and Systems (IROS)}, 2017, pp. 3562--3568.

\bibitem{Whitman2020}
J.~Whitman, \emph{et~al.}, ``Modular robot design synthesis with deep reinforcement learning,'' in \emph{Proc. of the AAAI Conf. on Artificial Intelligence (AAAI)}, vol.~34, no.~06, 2020, pp. 10\,418--10\,425.

\bibitem{Park2021}
J.~H. Park and K.~H. Lee, ``Computational design of modular robots based on genetic algorithm and reinforcement learning,'' \emph{Symmetry}, vol.~13, no.~3, 2021.

\bibitem{Sims1994}
K.~Sims, ``Evolving virtual creatures,'' in \emph{Proc. of the Ann. Conf. on Computer Graphics and Interactive Techniques (SIGGRAPH)}, vol.~21, 1994, pp. 15--22.

\bibitem{Romiti2023}
E.~Romiti, \emph{et~al.}, ``An optimization study on modular reconfigurable robots: Finding the task-optimal design,'' in \emph{IEEE Int. Conf. on Automation Science and Engineering (CASE)}, 2023, pp. 1--8.

\bibitem{Kuelz2024}
J.~K\"{u}lz and M.~Althoff, ``Optimizing modular robot composition: A lexicographic genetic algorithm approach,'' in \emph{Proc. of the IEEE Int. Conf. on Robotics and Automation (ICRA)}, 2024, pp. 16\,752--16\,758.

\bibitem{Koos2010}
S.~Koos, J.-B. Mouret, and S.~Doncieux, ``Crossing the reality gap in evolutionary robotics by promoting transferable controllers,'' in \emph{Proc. of the Conf. on Genetic and Evolutionary Computation (GECCO)}, 2010, pp. 119--126.

\bibitem{Tobin2017}
J.~Tobin, \emph{et~al.}, ``Domain randomization for transferring deep neural networks from simulation to the real world,'' in \emph{Proc. of the IEEE/RSJ Int. Conf. on Intelligent Robots and Systems (IROS)}, 2017, pp. 23--30.

\bibitem{Mordatch2015}
I.~Mordatch, K.~Lowrey, and E.~Todorov, ``{Ensemble-CIO:} full-body dynamic motion planning that transfers to physical humanoids,'' in \emph{Proc. of the IEEE/RSJ Int. Conf. on Intelligent Robots and Systems (IROS)}, 2015, pp. 5307--5314.

\bibitem{Mehta2020}
B.~Mehta, \emph{et~al.}, ``Active domain randomization,'' in \emph{Proc. of the Conf. on Robot Learning (CoRL)}, L.~P. Kaelbling, D.~Kragic, and K.~Sugiura, Eds., vol. 100, 2020, pp. 1162--1176.

\bibitem{Andrychowicz2019}
M.~Andrychowicz, \emph{et~al.}, ``Learning dexterous in-hand manipulation,'' \emph{The International Journal of Robotics Research}, vol.~39, no.~1, pp. 3--20, 2019.

\bibitem{Fadini2022}
G.~Fadini, \emph{et~al.}, ``Simulation aided co-design for robust robot optimization,'' \emph{IEEE Robotics and Automation Letters}, vol.~7, no.~4, pp. 11\,306--11\,313, 2022.

\bibitem{Deb2002}
K.~Deb, \emph{et~al.}, ``A fast and elitist multiobjective genetic algorithm: {NSGA-II},'' \emph{IEEE Transactions on Evolutionary Computation}, vol.~6, no.~2, pp. 182--197, 2002.

\bibitem{Kuelz2023}
J.~K\"{u}lz, M.~Mayer, and M.~Althoff, ``{{Timor} {Python}}: A toolbox for industrial modular robotics,'' in \emph{Proc. of the IEEE/RSJ Int. Conf. on Intelligent Robots and Systems (IROS)}, 2023, pp. 424--431.

\bibitem{Mayer2024}
M.~Mayer, J.~K\"{u}lz, and M.~Althoff, ``{CoBRA}: A composable benchmark for robotics applications,'' in \emph{Proc. of the IEEE Int. Conf. on Robotics and Automation (ICRA)}, 2024, pp. 17\,665--17\,671.

\bibitem{Romiti2022}
E.~Romiti, \emph{et~al.}, ``Toward a plug-and-work reconfigurable cobot,'' \emph{Transactions on Mechatronics}, vol.~27, no.~5, pp. 2319--3231, 2022.

\bibitem{Macenski2020}
S.~Macenski, \emph{et~al.}, ``The {Marathon} 2: A navigation system,'' in \emph{Proc. of the IEEE/RSJ Int. Conf. on Intelligent Robots and Systems (IROS)}, 2020, pp. 2718--2725.

\bibitem{macenski2024smac}
S.~Macenski, M.~Booker, and J.~Wallace, ``Open-source, cost-aware kinematically feasible planning for mobile and surface robotics,'' arXiv:2401.13078, 2024.

\bibitem{Jurado2016}
S.~Garrido-Jurado, \emph{et~al.}, ``Generation of fiducial marker dictionaries using mixed integer linear programming,'' \emph{Pattern Recognition}, vol.~51, pp. 481--491, 2016.

\bibitem{Schroeder2016basics}
B.~Schr\"{o}der, \emph{Ordered Sets}.\hskip 1em plus 0.5em minus 0.4em\relax Birkh\"{a}user Cham, 2016, vol.~2, ch. 1, Basics, pp. 1--21.

\bibitem{Harzheim2005orders}
E.~Harzheim, \emph{Ordered Sets}.\hskip 1em plus 0.5em minus 0.4em\relax Springer, 2005, ch. 4, Products of orders, pp. 85--141.

\bibitem{Boyd2023convex}
S.~P. Boyd, \emph{Convex optimization}.\hskip 1em plus 0.5em minus 0.4em\relax Cambridge University Press, 2023, vol.~7, ch. 4, Convex optimization problems, pp. 141--227.

\bibitem{Coello2000}
C.~A. Coello, ``An updated survey of {GA}-based multiobjective optimization techniques,'' \emph{{ACM} Computing Surveys}, vol.~32, no.~2, pp. 109--143, 2000.

\end{thebibliography}

\appendix
\subsection{NSGA-II metrics}\label{appdx:nsga2_metrics}
\xhdr{Nondominated fronts}
The Pareto front $\sF^0 \subset \sF$ consists of the Pareto-optimal elements of a set $\sF$, i.e., all elements not dominated by any other element of $\sF$.
By removing all Pareto-optimal elements from a set, i.e., performing a reduction, we obtain a reduced set $\sF^{1} = \sF \setminus \sF^0$.
All elements on the Pareto-front of $\sF^{1}$ have a nondomination rank of one as they are one reduction operation away from being Pareto-optimal with respect to $\sF$.
This process can be repeated until every element of $\sF$ is assigned a nondomination rank.
The NSGA-II algorithm is based on the concept of nondomination fronts for optimization problems with equally important objectives~\cite{Deb2002}.
In our example, robot two has a nondomination rank of one.
All other robots have a nondomination rank of zero, as they are on the Pareto front of the set $\{\robot_1, \robot_2, \robot_3, \robot_4\}$.

\xhdr{Crowding distance}
Sometimes, many solutions for an optimization problem might be on the same nondomination front.
When performing selection within a genetic algorithm, in this scenario, a spread of solutions along the front is often preferred over random sampling.
The \ac{nsga2} algorithm introduces the crowding distance to systematically encode this preference.
For every individual, it is composed of the distance to the other individuals for every objective function.
A formal definition can be found in~\cite{Deb2002}.

\subsection{Task representation}\label{appdx:cobra}
The environment is represented as a set of obstacles $\obstacles \subset \R^3$.
A goal $\goal \in \goals$ is a tuple $(\pose, \tol)$ consisting of a desired pose $\pose$ and a tolerance $\tol$.
The pose defines a desired end-effector position $\destrans \in \mathbb{R}^3$ and orientation $\desrot \in SO(3)$.
We introduce the pose error between a desired pose $\pose$ and an actual end-effector pose $\eefpose$
\begin{align}
    \poseerr(\pose, \eefpose) = \left( \transerr, \roterr \right), && \transerr \in \mathbb{R}^3, \roterr \in SO(3) \, , \label{eqn:poseerr}
\end{align}
where $\transerr$ and $\roterr$ denote the translation and rotation offsets.

\subsection{Tolerances}\label{appdx:tolerances}
A tolerance $\tol: SE(3) \to \{\istrue, \isfalse\}$ maps the error term in~\eqref{eqn:poseerr} (here written as a tuple) to a binary classification that indicates whether the tolerance is met.
We introduce the axis-angle representation of a rotation according to Euler's rotation theorem:
Any rotation $\mR \in SO(3)$ can be represented by a unit axis $\vn(\mR)$, and a rotation angle
\begin{align}
    \axisangle(\mR) = \arccos \left( \frac{\trace{\mR} - 1}{2} \right) \, . \label{eq:axis_angle}
\end{align}
We introduce a general Euclidean tolerance $\euclideantol$ and the general axis tolerance $\axistol$ with 
tolerance axis $\vu$, where $\sum_i \evu_i = 1 \land \evu_i > 0$ and
numerical threshold $\boldsymbol{\eps}$, where $0 < \eps_i \ll 1$ as
\begin{align}
    \euclideantol(\poseerr) &=
    \begin{cases}
        \istrue& \textif \abs{\transerr} \leq \begin{bmatrix} x & y & z \end{bmatrix}^T \\
        \isfalse& \otherwise.
    \end{cases} \label{eq:eucliden_tolerance} \\
    \axistol(\poseerr) &=
    \begin{cases}
        \istrue&  \textif \axisangle(\roterr) \leq \tolangle \\
        & \qquad \lor \abs{\vn(\roterr) - \vu } \leq \boldsymbol{\eps} \\
        & \qquad \lor \abs{\vn(\roterr) + \vu } \leq \boldsymbol{\eps} \\
        \isfalse& \otherwise.
    \end{cases} \label{eq:axis_angle_tolerance}
\end{align}
Here, $\leq$ and $\abs{\cdot}$ denote element-wise order and absolute value.
The general Euclidean tolerance constrains the end effector within a box around the desired pose.
The axis tolerance is met for arbitrary rotations around axes $\pm \vu$ or small rotations of at most $\tolangle$ about any axis.
The parameter $\boldsymbol{\eps}$ accounts for the inaccuracy of numerical computations of the orientation error.
For all experiments, we used the Euclidean tolerance $\euclideantol$ with $x = y = 0.2 \mm$ and $z = 10 \mm$ and the axis tolerance $\axistol$ with $\vu = \begin{bmatrix}
    0 & 0 & 1
\end{bmatrix}^T$, $\tolangle = 2\deg$, and $\boldsymbol{\eps} = 10^{-3} \begin{bmatrix} 1 & 1 & 1 \end{bmatrix}^T$, that allow for arbitrary rotations around and small translations along the drill axis parallel to $z$ but enforce a precise tracking of the desired waypoints.

\begin{IEEEbiography}[{\includegraphics[width=1in,height=1.25in,clip,keepaspectratio]{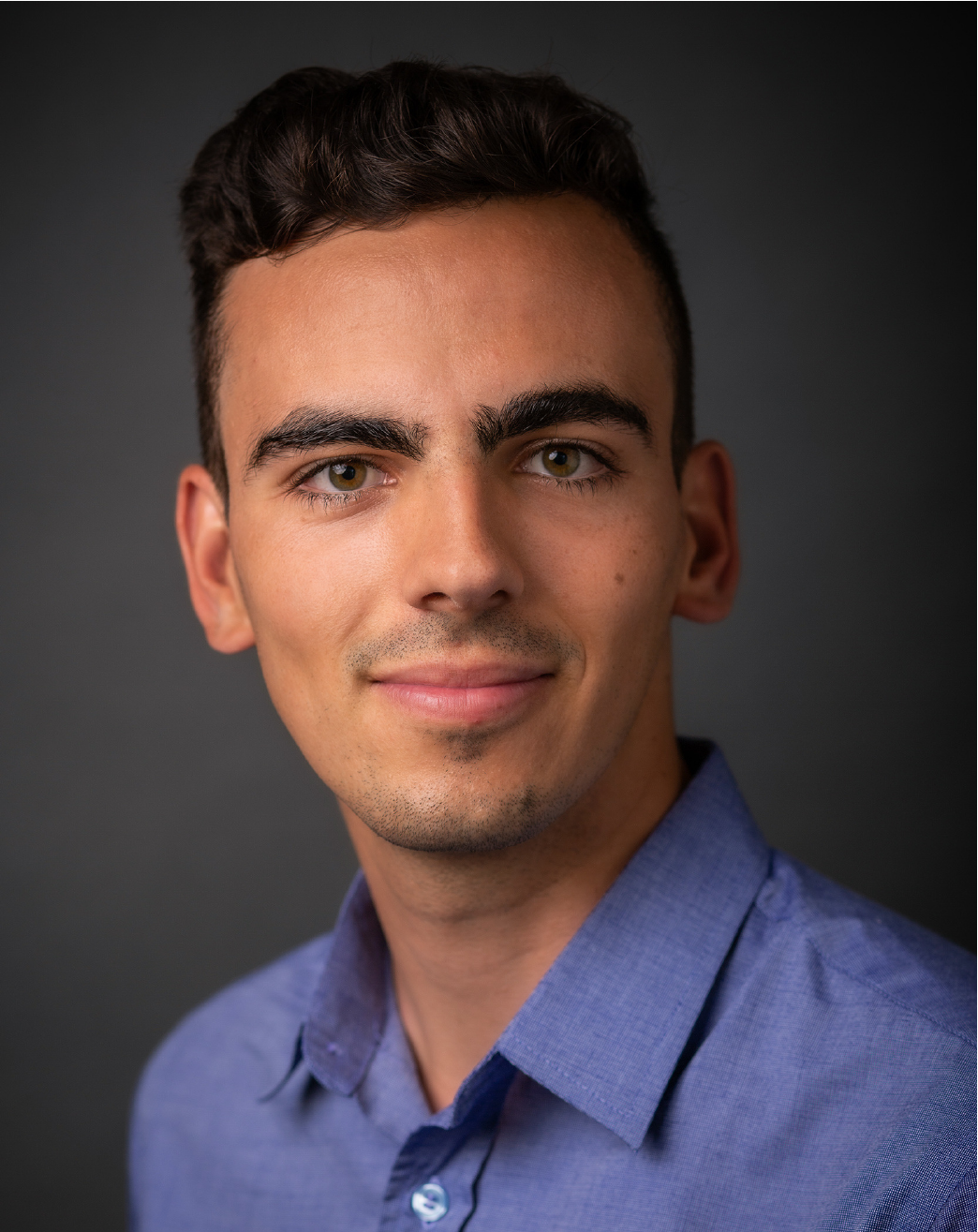}}]
	{Jonathan K\"ulz} received a bachelor's degree in mechatronics and information technology in 2017 from the Karlsruhe Institute of Technology, Germany, and a master's degree in robotics, cognition, and intelligence, in 2021, from the Technical University of Munich, Germany, where he is currently working toward the Ph.D. degree in computer science.
    His research interests include morphology optimization for modular robots, robot kinematics, robot co-design, and industrial automation.
\end{IEEEbiography}
\begin{IEEEbiography}[{\includegraphics[width=1in,height=1.25in,clip,keepaspectratio]{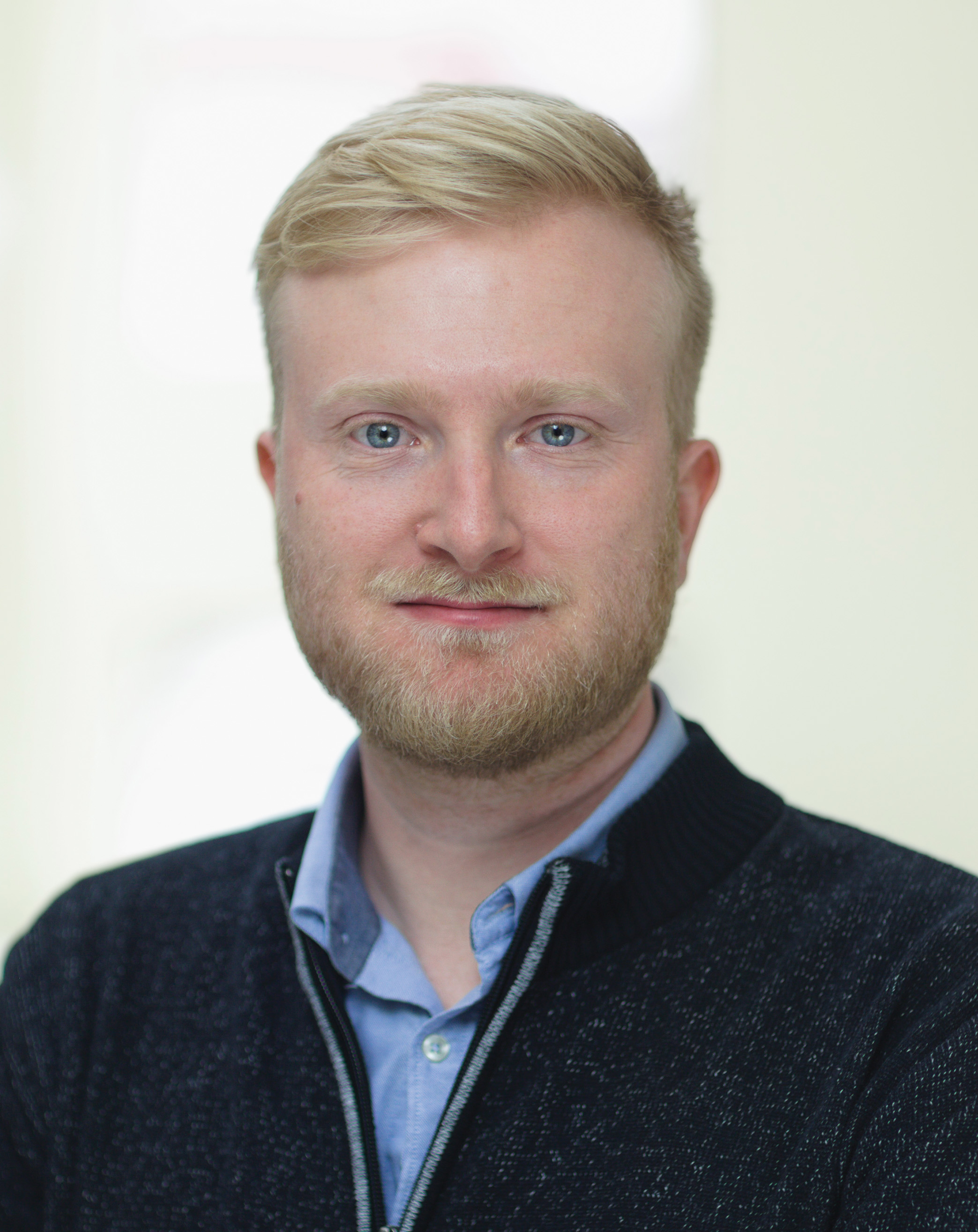}}]{Michael Terzer} received the bachelor's degree in mechatronics from the University of Innsbruck, Austria in 2015 and the master's degree in mechatronics and smart technologies from the Management Center Innsbruck, Austria in 2017. He is currently a Researcher and Team Leader within the group of robotics and intelligent systems engineering at Fraunhofer Italia Research, Bolzano, Italy. His research interests include robot mission planning, design and control of mechatronic and robotic systems, field robotics and mobile robot manipulators.
\end{IEEEbiography}
\begin{IEEEbiography}[{\includegraphics[width=1in,height=1.25in,clip,keepaspectratio]{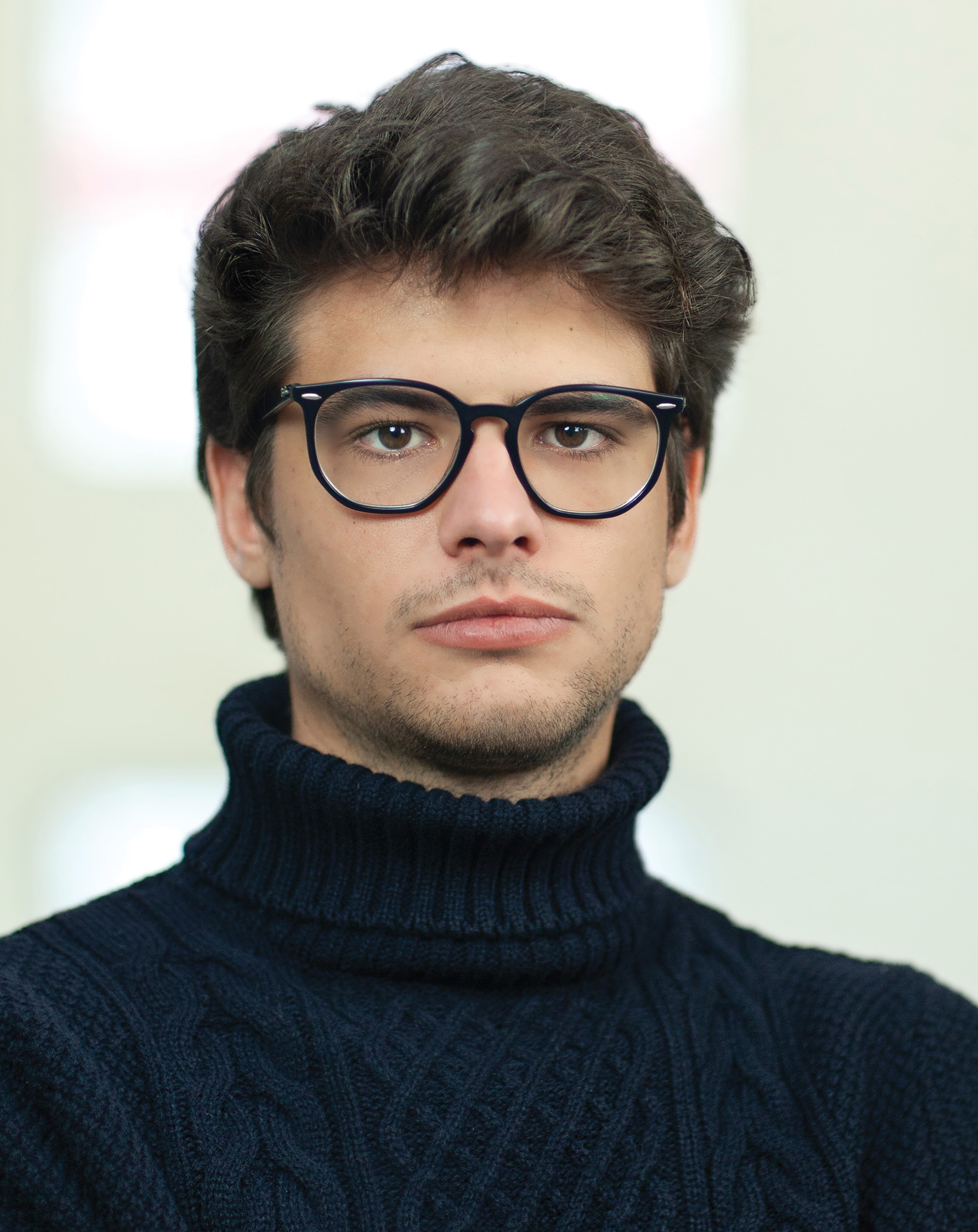}}]{Marco Magri} received his
bachelor’s degree in automation engineering in 2019, and his master’s degree in automation engineering in 2021, both from
University of Bologna, Italy. He is currently a Researcher at Fraunhofer
Italia Research, Italy. His research interests include robotics, computer vision and artificial intelligence.
\end{IEEEbiography}
\begin{IEEEbiography} [{\includegraphics[width=1in,height=1.25in,clip,keepaspectratio]{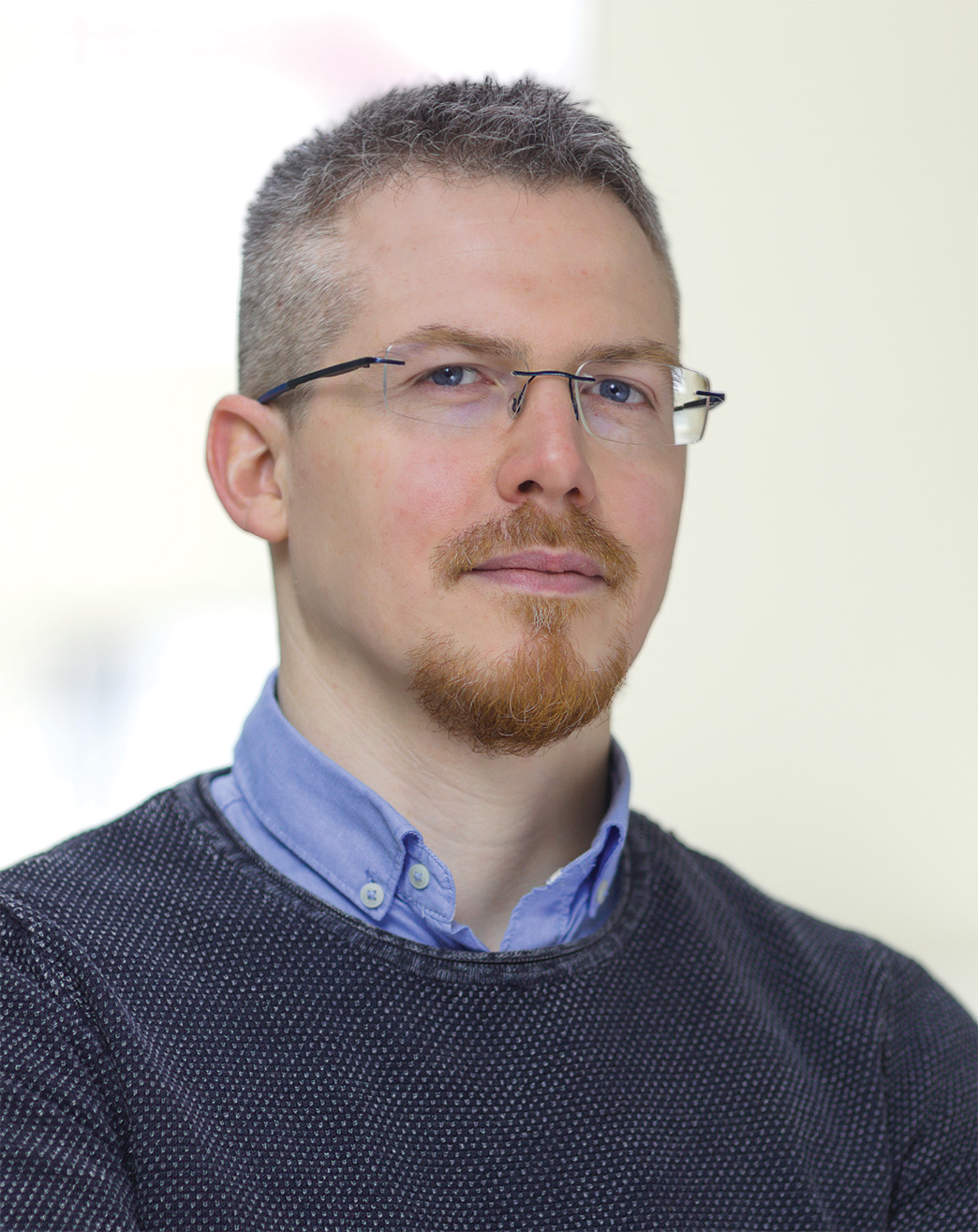}}]{Andrea Giusti} received the bachelor's degree in telecommunications engineering and the master's degree in mechatronic engineering from the University of Trento, Italy, in 2010 and 2013, respectively, and the Ph.D. degree in robotics from the Technical University of Munich, Germany, in 2018. He is currently a Researcher and Head of robotics and intelligent systems engineering with Fraunhofer Italia Research, Bolzano, Italy. His research interests include modelling and control of robotic and mechatronic systems, modular and reconfigurable robots, and human-robot collaboration.
\end{IEEEbiography}
    \begin{IEEEbiography}[{\includegraphics[width=1in,height=1.25in,clip,keepaspectratio]{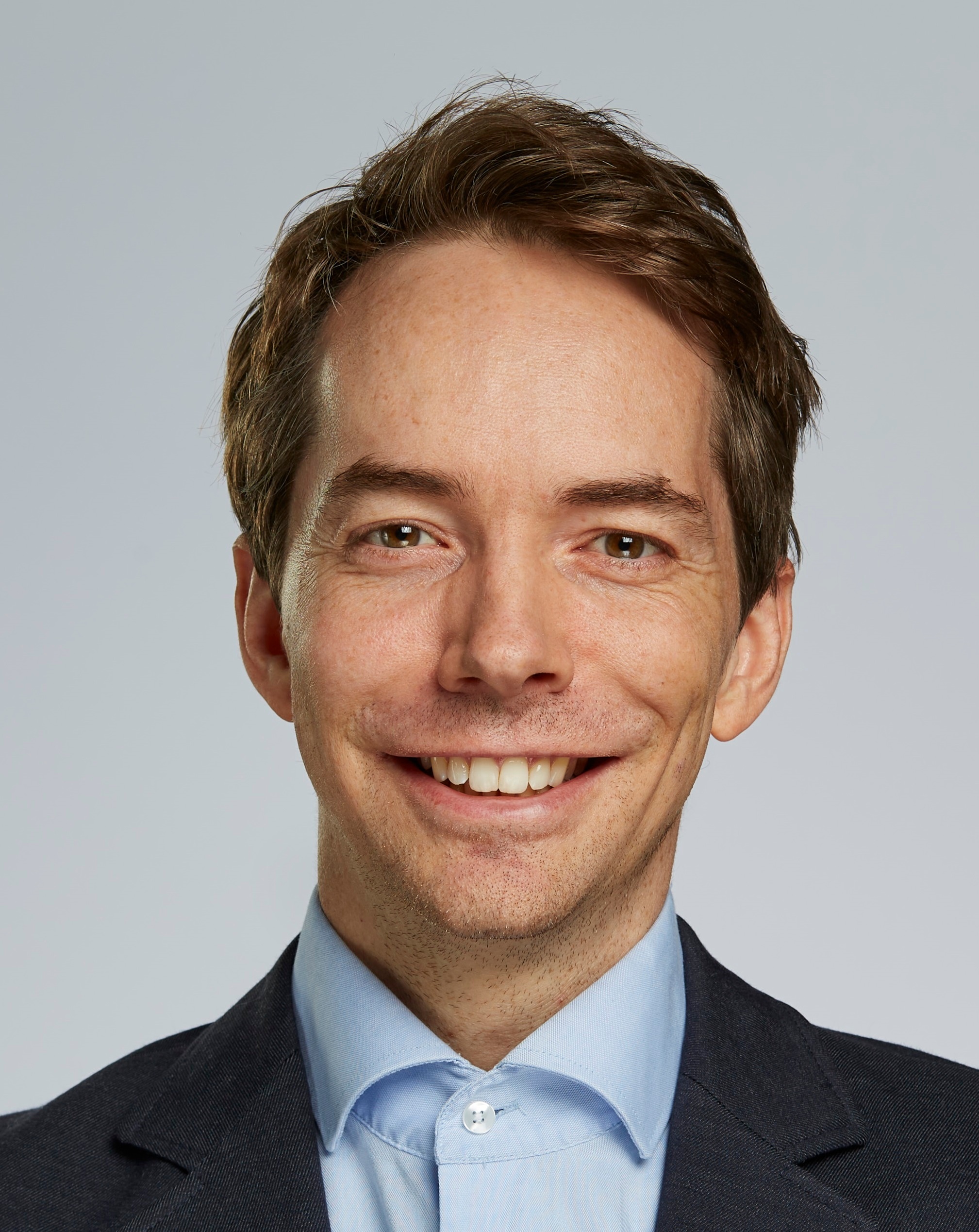}}]
    	{Matthias Althoff} is an associate professor in computer science at the Technical University of Munich, Germany. He received his diploma engineering degree in Mechanical
    	Engineering in 2005, and his Ph.D. degree in Electrical Engineering in
    	2010, both from the Technical University of Munich, Germany.
    	From 2010 to 2012 he was a postdoctoral researcher at Carnegie Mellon University,
    	Pittsburgh, USA, and from 2012 to 2013 an assistant professor at Technische Universit\"at Ilmenau, Germany. His research interests include formal verification of continuous and hybrid systems, reachability analysis, planning algorithms, safe machine learning, automated vehicles, and robotics.
    \end{IEEEbiography}
\vfill

\end{document}